\documentclass[twocolumn]{IEEEtran}
\usepackage[T1]{fontenc}
\usepackage[utf8x]{inputenc}
\usepackage{color}
\usepackage{pifont}
\usepackage{float}
\usepackage{units}
\usepackage{mathrsfs}
\usepackage{url}
\usepackage{amsmath}
\usepackage{amsthm}
\usepackage{amssymb}
\usepackage{graphicx}
\usepackage[unicode=true,
 bookmarks=true,bookmarksnumbered=true,bookmarksopen=true,bookmarksopenlevel=1,
 breaklinks=false,pdfborder={0 0 0},pdfborderstyle={},backref=false,colorlinks=false]
 {hyperref}
\hypersetup{pdftitle={Adaptive Constrained Kinematic Control using Partial or Complete Task-Space Measurements},
 pdfauthor={Murilo M. Marinho and Bruno V. Adorno},
 pdfpagelayout=OneColumn, pdfnewwindow=true, pdfstartview=XYZ, plainpages=false}

\makeatletter

\providecommand{\tabularnewline}{\\}
\floatstyle{ruled}
\newfloat{algorithm}{tbp}{loa}
\providecommand{\algorithmname}{Algorithm}
\floatname{algorithm}{\protect\algorithmname}

\let\oldforeign@language\foreign@language
\DeclareRobustCommand{\foreign@language}[1]{%
  \lowercase{\oldforeign@language{#1}}}
\theoremstyle{plain}
\newtheorem{thm}{\protect\theoremname}
\theoremstyle{remark}
\newtheorem{rem}[thm]{\protect\remarkname}
\theoremstyle{plain}
\newtheorem{lem}[thm]{\protect\lemmaname}

\usepackage[caption=false,font=footnotesize]{subfig}

\usepackage{graphicx}
\usepackage{import}

\usepackage{balance}

\usepackage{resizegather}

\usepackage{algpseudocode,algorithm}

\usepackage{cite}

\usepackage[below]{placeins}
\usepackage{afterpage}

\makeatother

\providecommand{\lemmaname}{Lemma}
\providecommand{\remarkname}{Remark}
\providecommand{\theoremname}{Theorem}

\begin{document}
\global\long\def\dq#1{\underline{\boldsymbol{#1}}}%

\global\long\def\quat#1{\boldsymbol{#1}}%

\global\long\def\mymatrix#1{\boldsymbol{#1}}%

\global\long\def\myvec#1{\boldsymbol{#1}}%

\global\long\def\mapvec#1{\boldsymbol{#1}}%

\global\long\def\dualvector#1{\underline{\boldsymbol{#1}}}%

\global\long\def\dual{\varepsilon}%

\global\long\def\dotproduct#1{\langle#1\rangle}%

\global\long\def\norm#1{\left\Vert #1\right\Vert }%

\global\long\def\mydual#1{\underline{#1}}%

\global\long\def\hamilton#1#2{\overset{#1}{\operatorname{\mymatrix H}}\left(#2\right)}%

\global\long\def\hamiquat#1#2{\overset{#1}{\operatorname{\mymatrix H}}_{4}\left(#2\right)}%

\global\long\def\hami#1{\overset{#1}{\operatorname{\mymatrix H}}}%

\global\long\def\tplus{\dq{{\cal T}}}%

\global\long\def\getp#1{\operatorname{\mathcal{P}}\left(#1\right)}%

\global\long\def\getd#1{\operatorname{\mathcal{D}}\left(#1\right)}%

\global\long\def\swap#1{\text{swap}\{#1\}}%

\global\long\def\imi{\hat{\imath}}%

\global\long\def\imj{\hat{\jmath}}%

\global\long\def\imk{\hat{k}}%

\global\long\def\real#1{\operatorname{\mathrm{Re}}\left(#1\right)}%

\global\long\def\imag#1{\operatorname{\mathrm{Im}}\left(#1\right)}%

\global\long\def\imvec{\boldsymbol{\imath}}%

\global\long\def\vector{\operatorname{vec}}%

\global\long\def\mathpzc#1{\fontmathpzc{#1}}%

\global\long\def\cost#1#2{\underset{\text{#2}}{\operatorname{\text{cost}}}\left(\ensuremath{#1}\right)}%

\global\long\def\diag#1{\operatorname{diag}\left(#1\right)}%

\global\long\def\frame#1{\mathcal{F}_{#1}}%

\global\long\def\ad#1#2{\text{Ad}\left(#1\right)#2}%

\global\long\def\totalderivative#1#2{\frac{\partial\left(#1\right)}{\partial#2}}%

\global\long\def\veceight#1{\operatorname{vec}_{8}#1}%

\global\long\def\unitdqspace{\dq{\mathcal{S}}}%

\global\long\def\unitquatspace{\mathbb{S}^{3}}%

\global\long\def\hp{\mathbb{H}_{p}}%

\global\long\def\spin{\text{Spin}(3)}%

\global\long\def\spinr{\text{Spin}(3){\ltimes}\mathbb{R}^{3}}%

\global\long\def\estimated#1{\hat{#1}}%

\global\long\def\jointspace{\mathcal{Q}}%

\global\long\def\parameterspace{\mathcal{A}}%

\global\long\def\taskspace{\mathscr{T}}%

\global\long\def\measurespace{\mathcal{Y}}%

\global\long\def\restrictedset#1{#1_{\mathrm{r}}}%

\global\long\def\argminimone#1#2#3#4{\begin{aligned}#1\:  &  \underset{#2}{\arg\!\min}  &   &  #3\\
  &  \text{subject to}  &   &  #4 
\end{aligned}
 }%

\global\long\def\minimtwo#1#2#3#4{ \begin{aligned} &  \underset{#1}{\min}  &   &  #2 \\
  &  \text{subject to}  &   &  #3 \\
  &   &   &  #4 
\end{aligned}
 }%

\global\long\def\minimone#1#2#3{ \begin{aligned} &  \underset{#1}{\min}  &   &  #2 \\
  &  \text{subject to}  &   &  #3 
\end{aligned}
 }%

\global\long\def\argminimtwo#1#2#3#4#5{ \begin{aligned}#1\:  &  \underset{#2}{\arg\!\min}  &   &  #3 \\
  &  \text{subject to}  &   &  #4\\
  &   &   &  #5 
\end{aligned}
 }%

\global\long\def\argmaximtwo#1#2#3#4#5{ \begin{aligned}#1\:  &  \underset{#2}{\arg\!\max}  &   &  #3 \\
  &  \text{subject to}  &   &  #4 \\
  &   &   &  #5 
\end{aligned}
 }%

\global\long\def\parametrictaskJcomplement{\text{parametric task-Jacobian projector}}%

\global\long\def\estimatedtaskerror{\text{\ensuremath{\breve{\myvec x}}}}%

\global\long\def\realtaskerror{\text{\ensuremath{\tilde{\myvec x}}}}%

\global\long\def\measurementerror{\text{\ensuremath{\tilde{\myvec y}}}}%

\global\long\def\bq{\myvec b_{q}}%

\global\long\def\bahat{\myvec b_{\estimated a}}%

\title{Adaptive Constrained Kinematic Control using Partial or Complete Task-Space
Measurements}
\author{Murilo~M.~Marinho,~\IEEEmembership{Member,~IEEE} and Bruno~V.~Adorno,~\IEEEmembership{Senior Member,~IEEE}\thanks{This
work was supported in part by JSPS KAKENHI Grant Number JP19K14935
and in part by the ImPACT Program of Council for Science, Technology
and Innovation Grant Number 2015-PM15-11-01. \emph{(Corresponding
author:} Murilo M. Marinho.)}\thanks{Murilo M. Marinho is with the
Department of Mechanical Engineering, the University of Tokyo, Tokyo,
Japan. \texttt{Email:murilo@g.ecc.u-tokyo.ac.jp}. }\thanks{Bruno
V. Adorno is with the Department of Electrical and Electronic Engineering,
The University of Manchester, Sackville Street, Manchester, M13 9PL,
United Kingdom . \texttt{Email: bruno.adorno@manchester.ac.uk}.}}
\markboth{Accepted on the IEEE Transactions on Robotics.}{M. M. Marinho and B. V. Adorno: Adaptive Constrained Kinematic Control
using Partial or Complete Task-Space Measurements}
\maketitle
\begin{abstract}
Recent advancements in constrained kinematic control make it an attractive
strategy for controlling robots with arbitrary geometry in challenging
tasks. Most current works assume that the robot kinematic model is
precise enough for the task at hand. However, with increasing demands
and safety requirements in robotic applications, there is a need for
a controller that compensates online for kinematic inaccuracies. We
propose an adaptive constrained kinematic control strategy based on
quadratic programming, which uses partial or complete task-space measurements
to compensate online for calibration errors. Our method is validated
in experiments that show increased accuracy and safety compared to
a state-of-the-art kinematic control strategy.
\end{abstract}

\begin{IEEEkeywords}
Robust/Adaptive Control, Kinematics, Optimization and Optimal Control.
\end{IEEEkeywords}

\section{Introduction}

\IEEEPARstart{K}{inematic control} has been applied effectively to
a myriad of tasks that use velocity-actuated robots with distinct
geometry, such as manipulator robots, mobile robots, and humanoid
robots.

Kinematic control strategies use a model derived from the geometric
parameters of the robotic system \cite{Siciliano2009}. The accuracy\footnote{Accuracy is a measure of how close a robot is able to move its end
effector to a desired point in its task space.} of the control strategy is directly related to the accuracy of the
robot's geometric parameters, such as link length for manipulator
robots, wheel radius for mobile robots, and so on. Moreover, in applications
that require cooperation between robots and/or between robots and
humans, the geometric parameters can also include the relative pose
between the reference frame of each robot and other entities in the
workspace.

In applications in which accuracy is not a major requirement, using
the parameters defined in the robot's design plans (schematics, computer-assisted
design files, etc.) is accurate enough. For some applications that
require a finer degree of accuracy, it is common practice to perform
a calibration process before executing the task to obtain a more reliable
estimate of the robot parameters \cite{Lenza1988,mooring1991fundamentals,Motta2001,Zhang2017}.
A calibration process that happens before the task execution is hereby
called an \emph{offline} calibration strategy.

In this work, our main motivations are applications in which offline
calibration is impractical and/or insufficient, and task-space constraints
are necessary. A few examples include medical applications in constrained
workspaces, in which the robot might have to be repositioned to be
used in different steps of the surgery \cite{marinho2019integration};
reconfigurable robots, for which the kinematics may change according
to novel configurations, especially when the coupling between different
robot parts within a new configuration is not precise enough; or even
assistant mobile manipulators handling different loosely grasped tools.
In those cases, it is not practical (or even impossible) to temporarily
stop the task to perform a time-consuming calibration procedure.

Moreover, in realistic applications, sensors that provide \emph{complete}
task-space measurements might not be available, especially when the
workspace is constrained. For instance, in robot-assisted surgery
for endonasal \cite{marinho2019integration} or neonate procedures
\cite{Marinho2021}, the constrained workspace prevents the use of
extra sensors. Also, having a reliable estimate of the robotic instruments'
tip pose just from the endoscopic camera is challenging. A more realistic
approach would be to use the information of the instruments' shaft
centerline \cite{Wang2018,yoshimura2020single}.

To further enable using robots in realistic and challenging scenarios,
our interest lies in integrating an \emph{online} calibration strategy
with constrained kinematic control to make the best of either \emph{partial}
or complete task-space measurements. In this sense, online refers
to calibration being performed \emph{at the same time} as the task
is being executed by the robot, resulting in an adaptive controller.

\subsection{Related works\label{subsec:Related-works}}

Constrained kinematic (and dynamic) control has received a lot of
attention and has been extended to take into account task inequalities
\cite{kanoun2011kinematic}, provide a fast hierarchical solution
\cite{Escande2014}, prevent collision with static and moving obstacles
\cite{Marinho2019}, and prevent self collisions \cite{Quiroz_Omana_2019}.
All these works suppose that the robot model is precise enough after
an offline calibration is performed.

Offline calibration strategies rely on having high-precision measuring
equipment in a highly controlled setting \cite{Lenza1988,mooring1991fundamentals,Motta2001,Zhang2017}.
In those approaches, the robotic system is placed in the calibration
setup that might be composed of cameras and markers, coordinate-measurement
machines, etc. Data are obtained from moving the robot around the
workspace and measuring end-effector poses by using markers with known
geometry or with respect to reference objects in the workspace. The
kinematic parameters are obtained from an optimization algorithm that
minimizes the error between the measured data (e.g., pose, position,
orientation) and the corresponding data obtained from the estimated
parameters.

Inspired by those offline techniques, recent works \cite{Gharaaty2018,Yu2018}
have integrated high-precision sensors in the robot's workspace in
industrial settings, which allows for the periodic update of the robot
parameters. However, those works do not provide updates while the
task is being performed.

Recent works have explored the optimal path to calibrate the robot's
dynamic and geometric parameters in an offline fashion \cite{Bonnet2018,Katsumata2019}.
Those works use constrained quadratic optimization strategies to minimize
the time required for offline calibration by proving a ``sufficiently
rich'' trajectory \cite{Morgan1977,slotine1991applied}.

To the best of our knowledge, one of the best techniques to the target
application of this work is adaptive control because it compensates
for inaccuracies in the robot model while the task is performed. For
instance, the dynamic parameters \cite{Slotine1987,Garofalo_2021}
and some kinematic parameters \cite{Cheah2006} of robots can be estimated
and adjusted online. However, there are two major drawbacks of most
existing adaptive controllers in this context: first, with very few
exceptions, most of them require linearity on the kinematic parameters
\cite{Lewis2004} because they rely on the so-called \emph{parameter
regression matrix}; second, they cannot consider task-space inequality
constraints. Most of the research making use of the parameter regression
matrix is in the context of grasping objects with uncertain weight
using dynamic control. In that case, there are efficient algorithms
to obtain the parameter regression matrix when, for instance, only
the inertial parameters are to be adapted \cite{Jing_Yuan}. A parameter
regressor can be found algebraically when the system is simple and
there are few parameters to be adapted \cite{slotine1991applied,Cheah_2004,Cheah2006}.
Based on the observations that it is not possible, in general, to
find linearity on the kinematic parameters, Marcucci et al. \cite{Marcucci2017}
proposed a technique to factorize the robot kinematics into a regressor
matrix and a vector of nonlinear functions of the uncertain parameters.
Then an adaptive control law that updates the estimate of these nonlinear
functions of the unknown parameters is proposed. Nonetheless, in addition
to assuming that the model is linearly parameterizable and that the
robot Jacobian is invertible, they do not account for constraints,
and the controller requires complete task-space measurements. Lastly,
the proof of closed-loop stability in existing works on adaptive control
is not trivially extensible to constrained kinematic control, especially
when inequality constraints are considered.

Data-driven approaches such as reinforcement learning \cite{Osa2018a}
aim to find a suitable reward function to perform a task and has been
idealized to take into account autonomous systems with arbitrary geometry.
The technique sought in this work would be complementary to research
efforts in reinforcement learning. This is because we aim for an adaptive
constrained controller that guarantees safety and adapts the robot's
uncertain model online according to available measurements. Therefore,
it could also be used to improve sim-to-real transfer \cite{Okada2021,Fischer2021}
by adapting the learned robot's model online. Similarly, many trajectory
planning approaches \cite{Finean2021,Hu2021} could directly benefit
from such controller for the same reasons. Part of the planning can
be performed offline and the constrained adaptive constrained controller
can allow the transfer of this planning to real scenarios.

Visual servoing is a technique that relies on cameras to control
the motion of robotic systems. Most classic techniques in visual servoing
require the system's Jacobian to be invertible (at least those with
stability proof), need a continuous stream of reliable camera readings,
and do not address workspace constraints. Recent approaches \cite{Wu_2022}
take advantage of data-driven algorithms to overcome one or more of
those limitations. For instance, recent works address visual servoing
under temporary occlusions \cite{Shi_2019} or tackle configuration-space
constraints by using a data-driven strategy to learn the robot's model
\cite{Zhang_2018}. Related data-driven approaches allow for impressive
performance in robot grasping in low-risk and low-accuracy scenarios
\cite{Sadeghi_2018}. In contrast with these existing approaches
in visual servoing, we are interested in making use of \emph{any}
sensor (not only cameras) that can provide a possibly discontinuous
stream of data to update the robot model in scenarios that might demand
high-accuracy, with nonlinear workspace constraints, and without imposing
any requirement on the Jacobian, such as invertibility\@.

Also, it is important to mention that this work does not aim to address
hand/eye calibration of cameras, which has been extensively studied
in the literature \cite{Tsai1989,Wang2018,Zhong2020}. In the current
study, we address task-space measurements only, and the estimation
of a camera's intrinsic and extrinsic parameters are out of this paper's
scope.

\subsection{Statement of contributions}

In this work, we build upon our constrained kinematic controller \cite{Marinho2019}
and propose an adaptive constrained kinematic control strategy that
makes use of partial or complete task-space measurements while taking
into account both equality and inequality constraints that are linear
in the control inputs. To account for nonlinear constraints in both
task-space variables and robot configurations, we extend the vector
field inequalities (VFIs) \cite{Marinho2019} framework to our adaptive
controller, such that those nonlinear constraints are transformed
into linear differential inequalities in the control inputs.

Differently from existing approaches, our strategy does not demand
a parameter regression matrix, requiring only differentiability of
the forward kinematics with respect to the estimated parameters. Using
Lyapunov stability theory, we show that the closed-loop system is
stable in the sense that the error between the estimated end-effector
pose and the desired pose is always non-increasing. We perform two
experiments. In the first experiment, we perform a thorough experimental
evaluation of the proposed methodology, in which we vary the amount
of information available about the task-space. In the second experiment,
we evaluate the controller in a scenario requiring multiple nonlinear
task-space constraints.

\section{Problem Definition\label{sec:Problem-Definition}}

Before describing the proposed adaptive control strategy, this section
introduces the online calibration problem with (partial) measurements,
in general terms. Also, we present the mathematical notation used
in this paper, which is summarized in Table~\ref{tab:notation}.

\begin{table}[tbh]
\caption{\label{tab:notation}Mathematical notation.}

\noindent\resizebox{1.0\columnwidth}{!}{%
\begin{centering}
\begin{tabular}{c|l}
\hline 
Notation & Definition\tabularnewline
\hline 
\hline 
$\restrictedset{\jointspace}\subseteq\jointspace\subseteq\mathbb{R}^{n}$ & Restricted configuration space and configuration space of $n$ dimensions.\tabularnewline
$\restrictedset{\parameterspace}\subseteq\parameterspace\subseteq\mathbb{R}^{p}$ & Restricted parameter space and parameter space of $p$ dimensions.\tabularnewline
$\restrictedset{\taskspace}\subseteq\taskspace\subseteq\mathbb{R}^{m}$ & Restricted task-space and task-space of $m$ dimensions.\tabularnewline
$\measurespace\subseteq\mathbb{R}^{r}$ & Measurement space of $r$ dimensions, with $r\leq m$.\tabularnewline
$\overline{\measurespace}\subseteq\mathbb{R}^{\bar{r}}$ & Space of unmeasured variables of $\bar{r}$ dimensions.\tabularnewline
$s$ & Number of constraints dependent on both the control inputs and the
adaptation signal.\tabularnewline
$s_{q}$ & Number of constraints dependent only on the control inputs.\tabularnewline
$s_{\estimated a}$ & Number of constraints dependent only on the adaptation signal. \tabularnewline
$\myvec x\in\taskspace,\quat y\in\measurespace$ & Real task-space and measurement-space vectors, respectively.\tabularnewline
$\estimated{\myvec x}\in\taskspace,\estimated{\myvec y}\in\measurespace$ & Estimated task-space and measurement-space vectors, respectively.\tabularnewline
$\estimatedtaskerror\in\taskspace$ & Error between the estimated ($\hat{\myvec x}$) and desired ($\myvec x_{d}$)
task-space vectors.\tabularnewline
$\tilde{\myvec y}\in\measurespace$ & Error between the estimated ($\estimated{\myvec y}$) and real ($\myvec y$)
measure-space vectors.\tabularnewline
\hline 
\end{tabular}
\par\end{centering}
}
\end{table}

\subsection{Uncertain kinematic model}

Consider a velocity (or position) actuated robotic system\footnote{The robotic system can be comprised of any number of individual robots.}
with $n$ degrees-of-freedom (DoF) and configuration vector given
by $\myvec q\triangleq\myvec q\left(t\right)\in\restrictedset{\jointspace}$,
in which the restricted configuration space is given by
\begin{align*}
\restrictedset{\jointspace} & \triangleq\left\{ \myvec q\in\jointspace:\myvec q_{\min}\preceq\myvec q\preceq\myvec q_{\max}\text{ and }\myvec q_{\min},\myvec q_{\max}\in\jointspace\subseteq\mathbb{R}^{n}\right\} .
\end{align*}
 Let the \emph{$p$ real} parameters of the robot kinematic model,
which are impossible to measure directly, be $\myvec a\in\restrictedset{\parameterspace}$,
where
\[
\restrictedset{\parameterspace}\triangleq\left\{ \myvec a\in\parameterspace:\myvec a_{\min}\preceq\myvec a\preceq\myvec a_{\max}\text{ and }\myvec a_{\min},\myvec a_{\max}\in\parameterspace\subseteq\mathbb{R}^{p}\right\} .
\]

If all parameters and joint values were perfectly known, the forward
kinematics model (FKM) could be used to obtain the \emph{real} task-space
variables, $\myvec x\triangleq\myvec x\left(\myvec q,\myvec a\right)\in\taskspace\subseteq\mathbb{R}^{m}$.
The FKM is the nonlinear mapping
\begin{align}
\myvec f:\jointspace\times\parameterspace\rightarrow & \taskspace, & \left(\myvec q,\myvec a\right) & \mapsto\myvec x,\label{eq:fkm}
\end{align}
 and the restricted task space is
\[
\restrictedset{\taskspace}\triangleq\left\{ \myvec f\left(\myvec q,\myvec a\right)\in\taskspace:\myvec q\in\restrictedset{\jointspace},\,\myvec a\in\restrictedset{\parameterspace}\right\} \subseteq\taskspace.
\]

Let the \emph{estimated} parameters be $\estimated{\myvec a}\left(t\right)\triangleq\estimated{\myvec a}\in\restrictedset{\parameterspace}$.
The \emph{estimated} task-space vector $\estimated{\myvec x}\triangleq\estimated{\myvec x}\left(\myvec q,\estimated{\myvec a}\right)\in\restrictedset{\taskspace}$
is obtained as
\begin{equation}
\estimated{\myvec x}\triangleq\myvec f\left(\myvec q,\estimated{\myvec a}\right).\label{eq:estimated_fkm}
\end{equation}
The first time-derivative of \eqref{eq:estimated_fkm} is the \emph{estimated}
differential forward kinematic model (DFKM)
\begin{align}
\dot{\hat{\myvec x}} & =\underbrace{\frac{\partial\myvec f\left(\myvec q,\estimated{\myvec a}\right)}{\partial\myvec q}}_{\mymatrix J_{\hat{x},q}}\dot{\myvec q}+\underbrace{\frac{\partial\myvec f\left(\myvec q,\estimated{\myvec a}\right)}{\partial\hat{\myvec a}}}_{\mymatrix J_{\hat{x},\hat{a}}}\dot{\hat{\myvec a}},\label{eq:estimated_dfkm}
\end{align}
where $\mymatrix J_{\hat{x},q}\triangleq\mymatrix J_{\hat{x},q}\left(\myvec q,\estimated{\myvec a}\right)\in\mathbb{R}^{m\times n}$
is the estimated task Jacobian and $\mymatrix J_{\hat{x},\hat{a}}\triangleq\mymatrix J_{\hat{x},\hat{a}}\left(\myvec q,\estimated{\myvec a}\right)\in\mathbb{R}^{m\times p}$
is the estimated parametric Jacobian.

Let us consider a measurement system capable of measuring an $r$-dimensional
subset of the real task-space variables given by $\myvec y\triangleq\myvec y\left(t\right)\in\measurespace\subseteq\mathbb{R}^{r}$,
and the $\overline{r}$-unmeasured dimensions compose the space of
unmeasured variables\footnote{When both $\measurespace$ and $\overline{\measurespace}$ are parameterized
with minimal representations, $\bar{r}=m-r$. However, if nonminimal
representations are used, such that the number of independent coordinates
in $\taskspace$, $\measurespace$, and $\overline{\measurespace}$
are $m_{\mathrm{ind}}<m$, $r_{\mathrm{ind}}<r$, and $\bar{r}_{\mathrm{ind}}<\bar{r}$,
respectively, then $\bar{r}_{\mathrm{ind}}=m_{\mathrm{ind}}-r_{\mathrm{ind}}$
but $\bar{r}\neq m-r$.} $\overline{\measurespace}\subseteq\mathbb{R}^{\bar{r}}$. In addition,
let the estimated measure-space FKM be given by the surjection
\begin{align}
\myvec g & :\restrictedset{\jointspace}\times\restrictedset{\parameterspace}\twoheadrightarrow\measurespace, & \left(\myvec q,\estimated{\myvec a}\right) & \mapsto\hat{\myvec y},\label{eq:measure_space_estimated_fkm}
\end{align}
with DFKM
\begin{equation}
\dot{\hat{\myvec y}}=\underbrace{\frac{\partial\myvec g\left(\myvec q,\estimated{\myvec a}\right)}{\partial\myvec q}}_{\mymatrix J_{\hat{y},q}}\dot{\myvec q}+\underbrace{\frac{\partial\myvec g\left(\myvec q,\estimated{\myvec a}\right)}{\partial\estimated{\myvec a}}}_{\mymatrix J_{\hat{y},\hat{a}}}\dot{\hat{\myvec a}},\label{eq:measure_space_estimated_dfkm}
\end{equation}
where $\mymatrix J_{\hat{y},q}\triangleq\mymatrix J_{\hat{y},q}\left(\myvec q,\estimated{\myvec a}\right)\in\mathbb{R}^{r\times n}$
is the estimated measure-space task Jacobian and $\mymatrix J_{\hat{y},a}\triangleq\mymatrix J_{\hat{y},a}\left(\myvec q,\estimated{\myvec a}\right)\in\mathbb{R}^{r\times p}$
is the estimated measure-space parametric Jacobian.

\subsection{Problem statement and main assumptions\label{subsec:Problem-statement-and-assumptions}}

Given an external measurement system that provides partial or complete
measurements $\myvec y\in\mathcal{Y}$, drive the real task-space
variable $\myvec x(\myvec q(t),\myvec a(t))\in\restrictedset{\taskspace}$
as close as possible to a constant desired value $\myvec x_{d}\in\taskspace$.

We assume that:
\begin{enumerate}
\item Perfect measurements are obtained in the task space and the configuration
space. Any sensor that provides a partial or complete measurement
of the task-space can be used. 
\item All constraints are represented by differentiable functions and no
constraint is violated at $t=0$.
\item The control input $\myvec u_{q}=\dot{\myvec q}(t)=\myvec 0$ is a
feasible solution for all $t\geq0$, which means that the robot is
always able to stop.
\end{enumerate}

\section{Proposed control strategy\label{sec:Proposed-control-strategy}}

To control the robotic system while updating the robot kinematic parameters,
we propose a control strategy based on two independent quadratic\footnote{Linear-quadratic optimization problems with inequality constraints
cannot be solved analytically and are solved with numerical methods.} programming (QP) problems that are solved either in cascade or simultaneously
to generate each instantaneous control input.

\begin{figure*}[t]
\noindent \begin{centering}
\includegraphics[width=0.85\textwidth]{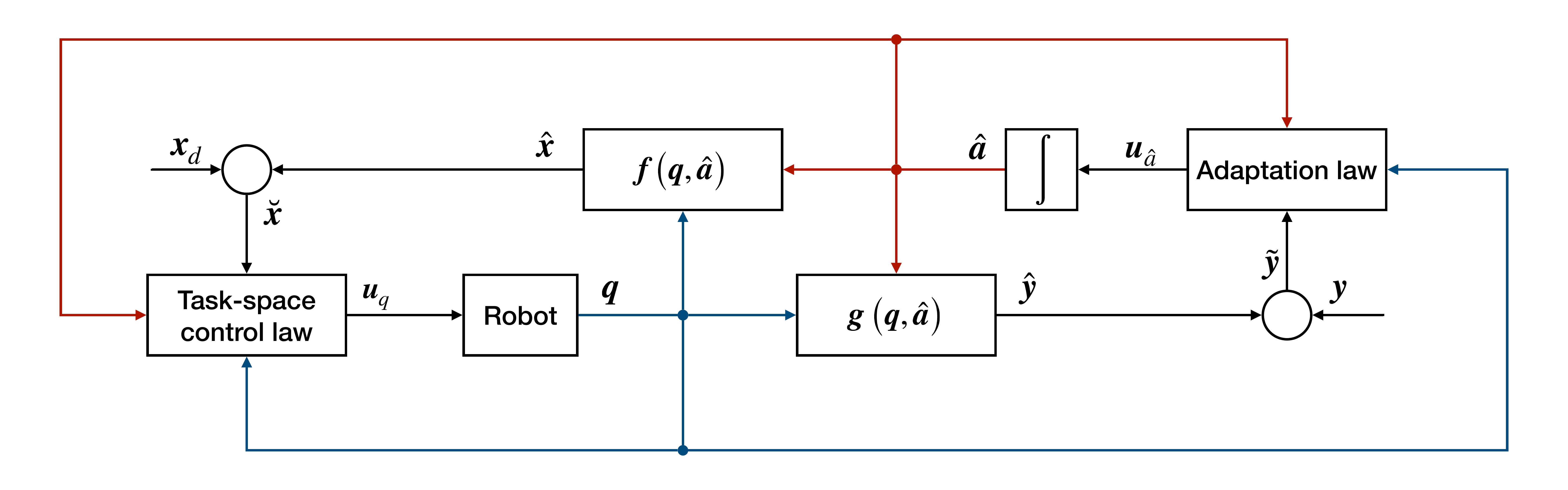}
\par\end{centering}
\caption{Block diagram of the adaptive constrained controller. The \emph{task-space
control law} \eqref{eq:problem_quadratic_constrained} aims to reduce
the estimated task-space error $\protect\estimatedtaskerror$ using
the current robot configuration vector $\protect\myvec q$, the estimated
parameter vector $\hat{\protect\myvec a}$, and the current desired
task-space value $\protect\myvec x_{d}$ to compute the next optimal
control input $\protect\myvec u_{q}$ sent to the robot. The \emph{adaptation
law} \eqref{eq:adaptive-constrained-law} aims to reduce the measurement
error $\protect\measurementerror$ using $\protect\myvec q$, $\hat{\protect\myvec a}$,
and the current task-space measurement $\protect\myvec y$ to compute
the optimal adaptation signal $\protect\myvec u_{\protect\estimated a}$.
The red lines indicate the feedback of the parameter vector $\hat{\protect\myvec a}$
whereas the blue lines indicate the feedback of the robot configuration
vector $\protect\myvec q$. \label{fig:Block-diagram-of}}
\end{figure*}

First, a task-space control law uses the estimated parameters, $\estimated{\myvec a}$,
and the desired task-space reference, $\myvec x_{d}$, to calculate
the optimal joint-velocity control input $\dot{\myvec q}\triangleq\quat u_{q}$,
which is sent to the robotic system. Then, an adaptive law generates
the optimal adaptation signal $\dot{\hat{\myvec a}}\triangleq\myvec u_{\hat{a}}$
using the current task-space measurement $\myvec y$ and robot configuration
$\myvec q$ to update the estimated kinematic parameters. The process
repeats until $\hat{\myvec x}(\myvec q(t),\hat{\myvec a}(t))$ becomes
as close as possible to $\myvec x_{d}$. The block diagram of the
control strategy is shown in Fig.~\ref{fig:Block-diagram-of}.

\subsection{Task-space control law}

The control input is obtained as
\begin{gather}
\argminimone{\quat u_{q}\in}{\dot{\myvec q}}{\norm{\mymatrix J_{\hat{x},q}\dot{\myvec q}+\eta_{q}\estimatedtaskerror}_{2}^{2}+\norm{\mymatrix{\Lambda}_{q}\dot{\myvec q}}_{2}^{2}}{\mymatrix B_{q}\dot{\myvec q}\preceq\bq}\label{eq:problem_quadratic_constrained}\\
\mymatrix W_{q}\dot{\myvec q}\preceq\myvec w_{q},\nonumber 
\end{gather}
where $\eta_{q}\in\left(0,\infty\right)$ is a proportional gain and
$\mymatrix{\Lambda}_{q}\in\mathbb{R}^{n\times n}$ is a positive definite
gain matrix, usually diagonal, that penalizes high joint-velocities.
In addition, $\mymatrix J_{\hat{x},q}$ is the estimated task Jacobian,
as defined in \eqref{eq:estimated_dfkm}, and $\estimatedtaskerror\triangleq\estimatedtaskerror\left(\hat{\myvec x}(t),\myvec x_{d}\right)$
is a suitable \emph{task error} that quantifies how close the estimated
task-space variable is to the desired setpoint. A common choice
is $\estimatedtaskerror\triangleq\estimated{\myvec x}-\myvec x_{d}$.
Furthermore, since $\hat{\myvec x}$ depends on $\estimated{\myvec a}$,
as shown in \eqref{eq:estimated_fkm}, the objective function in \eqref{eq:problem_quadratic_constrained}
also depends on the estimated parameters $\estimated{\myvec a}$.

If $\estimatedtaskerror=\myvec 0$, then $\quat u_{q}=\myvec 0$.
This is due to the fact that when $\estimatedtaskerror=\myvec 0$,
the objective function in \eqref{eq:problem_quadratic_constrained}
boils down to $\norm{\mymatrix J_{\hat{x},q}\dot{\myvec q}}_{2}^{2}+\norm{\mymatrix{\Lambda}_{q}\dot{\myvec q}}_{2}^{2}$,
which is minimized if and only if $\dot{\myvec q}=\myvec 0$. This
happens even when $\mymatrix J_{\hat{x},q}$ has a non trivial nullspace
because $\mymatrix{\Lambda}_{q}$ is positive definite, hence invertible.

\subsubsection*{Constraints}

The matrix $\mymatrix B_{q}\triangleq\mymatrix B_{q}\left(\myvec q,\hat{\myvec a}\right)\in\mathbb{R}^{s\times n}$
and $\myvec b\triangleq\myvec b\left(\myvec q,\hat{\myvec a}\right)\in\mathbb{R}^{s}$,
with $\bq\succeq\myvec 0$, define the $s$ linear (scalar) constraints
imposed on the control inputs that also depend on the estimated parameters.
Those constraints can be used to prevent self-collisions or collisions
with other objects in the workspace using VFIs \cite{Marinho2019},
as explained in Section~\ref{subsec:Vector-field-inequalities}.

The constraint $\mymatrix W_{q}\dot{\myvec q}\preceq\myvec w_{q}$,
where $\mymatrix W_{q}\triangleq\mymatrix W_{q}\left(\myvec q\right)\in\mathbb{R}^{s_{q}\times n}$
and $\myvec w\triangleq\myvec w\left(\myvec q\right)\in\mathbb{R}^{s_{q}}$,
is used to enforce the $s_{q}$ (scalar) constraints unrelated to
the estimated parameters. For instance, configuration velocity limits
$\myvec q_{v,\min}$, $\myvec q_{v,\max}$ can be trivially enforced
with
\begin{equation}
\begin{bmatrix}-\mymatrix I_{n\times n}\\
\mymatrix I_{n\times n}
\end{bmatrix}\dot{\myvec q}\leq\begin{bmatrix}-\myvec q_{v,\min}\\
\myvec q_{v,\max}
\end{bmatrix}.\label{eq:configurations_velocity_limits_constraint}
\end{equation}
Moreover, to define limits for the robot configurations, we start
by defining $\tilde{\myvec q}_{\max}(t)\triangleq\myvec q(t)-\myvec q_{\max}$
and $\tilde{\myvec q}_{\min}(t)\triangleq\myvec q(t)-\myvec q_{\min}$,
such that $\tilde{\myvec q}_{\max}(0)\preceq\myvec 0$ and $\tilde{\myvec q}_{\min}(0)\succeq\myvec 0$,
with the corresponding differential inequalities (notice that the
joint limits do not vary with time, i.e., $\dot{\myvec q}_{\min}=\dot{\myvec q}_{\max}=\myvec 0$)
\begin{align}
\dot{\tilde{\myvec q}}_{\min}\left(t\right)+\eta_{\tilde{q}}\tilde{\myvec q}_{\min}\left(t\right) & \succeq\myvec 0,\label{eq:configurations-lower-limit}\\
\dot{\tilde{\myvec q}}_{\max}\left(t\right)+\eta_{\tilde{q}}\tilde{\myvec q}_{\max}\left(t\right) & \preceq\myvec 0.\label{eq:configurations-upper-limit}
\end{align}
By Gronwall's Lemma \cite{gronwall}, \eqref{eq:configurations-upper-limit}
and \eqref{eq:configurations-lower-limit} ensure that $\tilde{\myvec q}_{\max}(t)\preceq e^{-\eta_{\tilde{q}}t}\tilde{\myvec q}_{\max}(0)\preceq\myvec 0$
and $\tilde{\myvec q}_{\min}(t)\succeq e^{-\eta_{\tilde{q}}t}\tilde{\myvec q}_{\min}(0)\succeq\myvec 0$
and can be rewritten as
\begin{align}
\begin{bmatrix}\mymatrix{-I}_{n\times n}\\
\mymatrix I_{n\times n}
\end{bmatrix}\dot{\myvec q} & \preceq-\eta_{\tilde{q}}\begin{bmatrix}-\tilde{\myvec q}_{\min}(t)\\
\tilde{\myvec q}_{\max}(t)
\end{bmatrix}.\label{eq:configurations_limits_constraint}
\end{align}
The configuration-only constraints can therefore be the composition
of \eqref{eq:configurations_velocity_limits_constraint} and \eqref{eq:configurations_limits_constraint}
\begin{equation}
\underbrace{\begin{bmatrix}-\mymatrix I_{n\times n}\\
\mymatrix I_{n\times n}\\
-\mymatrix I_{n\times n}\\
\mymatrix I_{n\times n}
\end{bmatrix}}_{\mymatrix W_{q}}\dot{\myvec q}\leq\underbrace{\begin{bmatrix}-\myvec q_{v,\min}\\
\myvec q_{v,\max}\\
\eta_{\tilde{q}}\tilde{\myvec q}_{\min}(t)\\
-\eta_{\tilde{q}}\tilde{\myvec q}_{\max}(t)
\end{bmatrix}}_{\myvec w_{q}}.\label{eq:configurations_only_constraints}
\end{equation}

\subsection{Adaptation law}

The adaptation law is given by
\begin{gather}
\argminimone{\myvec u_{\hat{a}}\in}{\dot{\hat{\myvec a}}}{\norm{\mymatrix J_{\estimated y,\estimated a}\dot{\hat{\myvec a}}+\eta_{\estimated a}\tilde{\myvec y}}_{2}^{2}+\norm{\mymatrix{\Lambda}_{\hat{a}}\dot{\hat{\myvec a}}}_{2}^{2}}{\begin{split}\mymatrix B_{\hat{a}}\dot{\hat{\myvec a}} & \preceq\bahat\\
\mymatrix W_{\estimated a}\dot{\estimated{\myvec a}} & \preceq\myvec w_{\estimated a}\\
\mymatrix N_{\hat{a}}\dot{\estimated{\myvec a}} & =\myvec 0\\
\estimatedtaskerror^{T}\mymatrix J_{\hat{x},\estimated a}\dot{\hat{\myvec a}} & \leq0,
\end{split}
}\label{eq:adaptive-constrained-law}
\end{gather}
where $\eta_{\estimated a}\in\left(0,\infty\right)$ is a proportional
gain and $\mymatrix{\Lambda}_{\hat{a}}\in\mathbb{R}^{p\times p}$
is a positive definite gain matrix, usually diagonal. In addition,
$\tilde{\myvec y}\triangleq\tilde{\myvec y}\left(\estimated{\myvec y}(t),\myvec y(t)\right)$
is the\emph{ estimation error}, that is, the error between the estimated
measure-space value and the real measure-space value. The estimation
error indirectly measures the parameter estimation error using the
measure-space estimated FKM. For example, $\tilde{\myvec y}\triangleq\estimated{\myvec y}-\myvec y$.
Nonetheless, any differentiable error function that satisfies 
\begin{equation}
\tilde{\myvec y}=\myvec 0\iff\estimated{\myvec y}=\myvec y\label{eq:error function measure space}
\end{equation}
is acceptable.

The parameters can be updated even when the task error is zero, as
long as the measurement error is not zero and the estimated task error
is not affected. This can be interpreted as moving the end effector
of a virtual robot made of estimated parameters $\estimated{\myvec a}$
in the direction of the real end-effector of the real robot made of
actual parameters $\myvec a$.\footnote{The attentive reader will notice that, when $\estimatedtaskerror=\myvec 0,$
the last constraint in \eqref{eq:adaptive-constrained-law} is innocuous
because it becomes $0\leq0$, which is always true regardless of $\dot{\hat{\myvec a}}$.
However, in this situation, the trajectories of the closed-loop system
have already converged to the invariant set determined by $\dot{V}(\estimatedtaskerror)=0$,
as shown in Section~\ref{subsec:Lyapunov-stability}, which means
that the norm of the task error cannot increase despite changes in
the estimated parameters $\hat{\myvec a}$.} The term $\norm{\mymatrix{\Lambda}_{\hat{a}}\dot{\estimated{\myvec a}}}_{2}^{2}$
can be used to scale parameters that use different units, to minimize
the detrimental effects that occur when the Jacobian is ill-conditioned,
and to guarantee that the parameters stop being updated when $\estimated{\myvec y}=\myvec y\iff\tilde{\myvec y}=\myvec 0$.
More specifically, $\tilde{\myvec y}=\myvec 0$ implies $\norm{\mymatrix J_{\hat{y},\hat{a}}\dot{\estimated{\myvec a}}}_{2}^{2}+\norm{\mymatrix{\Lambda}_{\hat{a}}\dot{\estimated{\myvec a}}}_{2}^{2}$,
where $\mymatrix{\Lambda}_{\hat{a}}$ is positive definite, which
is minimized if and only if $\dot{\estimated{\myvec a}}=\myvec 0$.

Given the definition of the measure-space FKM \eqref{eq:measure_space_estimated_fkm}
and \eqref{eq:error function measure space},
\[
\tilde{\myvec y}=\myvec 0\iff\estimated{\myvec y}=\myvec y\iff\myvec g\left(\myvec q,\estimated{\myvec a}\right)=\myvec g\left(\myvec q,\myvec a\right).
\]
However, as the measure-space FKM is in general not injective, we
cannot say that $\myvec a=\estimated{\myvec a}$ when $\tilde{\myvec y}=\myvec 0$.

The adaptation law has four linear (vector) constraints. The first
constraint is used to enforce task-space constraints that depend on
the parameters, in which $\mymatrix B_{\hat{a}}\triangleq\mymatrix B_{\hat{a}}(\mymatrix q,\hat{\mymatrix a})\in\mathbb{R}^{s\times p}$
and $\myvec b_{\hat{a}}\triangleq\myvec b_{\hat{a}}(\myvec q,\hat{\myvec a})\in\mathbb{R}^{s}$,
with $\myvec b_{\hat{a}}\succeq\myvec 0$, and $s$ is the number
of VFIs. That constraint is used to ensure that the first (vector)
constraint in \eqref{eq:problem_quadratic_constrained} is satisfied
even during the adaptation of parameters, and explained in detail
in Section~\ref{subsec:Vector-field-inequalities}. The second constraint
is used to enforce parameter bounds that are independent of the robot
configuration, analogously to \eqref{eq:configurations-lower-limit}–\eqref{eq:configurations_limits_constraint},
with $\mymatrix W_{\estimated a}\in\mathbb{R}^{s_{\estimated a}\times p}$
and $\myvec w_{\estimated a}\triangleq\myvec w_{\estimated a}\left(\estimated{\myvec a}\right)\in\mathbb{R}^{s_{\estimated a}}$.
The third constraint restricts the trajectories of the estimated parameters
to lie within an invariant set. This is to prevent disturbing unmeasured
variables, with $\mymatrix N_{\hat{a}}\triangleq\mymatrix N_{\hat{a}}\left(\myvec q,\estimated{\myvec a}\right)\in\mathbb{R}^{\bar{r}\times p}$
being the $\parametrictaskJcomplement$. The fourth constraint is
a Lyapunov constraint \cite{Goncalves2020}, explained in detail in
Section~\ref{subsec:Stability}.

\subsection{Vector field inequalities in an adaptive framework\label{subsec:Vector-field-inequalities}}

In both \eqref{eq:problem_quadratic_constrained} and \eqref{eq:adaptive-constrained-law},
we have constraints that are used to enforce VFIs \cite{Marinho2019};
namely, $\mymatrix B_{q}\dot{\myvec q}\preceq\bq$ and $\mymatrix B_{\hat{a}}\dot{\hat{\myvec a}}\preceq\bahat$.
The main idea of VFIs is as follows. Given $s$ differentiable functions
$h_{i}:\jointspace\times\parameterspace\to\mathbb{R}$, such that
$h_{i}\triangleq h_{i}(\myvec q(t),\hat{\myvec a}(t))$, we define
the feasible set as
\[
\mathscr{F}=\left\{ \left(\myvec q,\hat{\myvec a}\right)\in\jointspace\times\parameterspace:\bigwedge_{i=1}^{s}\bigl(h_{i}\left(\myvec q,\hat{\myvec a}\right)\leq0\bigr)\right\} .
\]
The $s$ VFIs are defined as $\dot{h}_{i}(t)+\text{\ensuremath{\eta_{v}}}h_{i}(t)\leq0$,
where $\eta_{v}\in(0,\infty)$. Let $\myvec h=\begin{bmatrix}h_{1} & \cdots & h_{s}\end{bmatrix}^{T}\in\mathbb{R}^{s}$,
then $\dot{\myvec h}=\begin{bmatrix}\mymatrix B_{q} & \mymatrix B_{\hat{a}}\end{bmatrix}\dot{\myvec v}$,
with $\myvec v\triangleq\begin{bmatrix}\myvec q^{T} & \hat{\myvec a}^{T}\end{bmatrix}^{T}$,
\begin{align*}
\mymatrix B_{q} & =\begin{bmatrix}\nabla_{q}(h_{1}) & \cdots & \nabla_{q}(h_{s})\end{bmatrix}^{T},\\
\mymatrix B_{\hat{a}} & =\begin{bmatrix}\nabla_{\hat{a}}(h_{1}) & \cdots & \nabla_{\hat{a}}(h_{s})\end{bmatrix}^{T},
\end{align*}
in which $\nabla_{q}(h_{i})\triangleq\begin{bmatrix}\partial h_{i}/\partial q_{1} & \cdots & \partial h_{i}/\partial q_{n}\end{bmatrix}^{T}$
and $\nabla_{\hat{a}}(h_{i})\triangleq\begin{bmatrix}\partial h_{i}/\partial\hat{a}_{1} & \cdots & \partial h_{i}/\partial\hat{a}_{p}\end{bmatrix}^{T}$.

From Gronwall's lemma, given the differential inequality $\dot{\myvec h}(t)+\eta_{v}\quat h(t)\preceq\myvec 0$,
if $\myvec h(0)\preceq\myvec 0$ when $t=0$, then $\myvec h(t)\preceq e^{-\eta_{v}t}\myvec h(0)\preceq\myvec 0$
for all $t\geq0$. Thus, we stack the $s$ scalar VFIs to obtain a
single differential inequality in vector form:
\begin{align}
\mymatrix B_{q}\dot{\myvec q}+\mymatrix B_{\hat{a}}\dot{\hat{\myvec a}} & \preceq-\eta_{v}\quat h(t).\label{eq:VFI}
\end{align}
To ensure \eqref{eq:VFI} by means of the first inequality in \eqref{eq:problem_quadratic_constrained}
and the first inequality in \eqref{eq:adaptive-constrained-law},
we define $\bq\triangleq-\eta_{v}\quat h(t)\left(1-\alpha\right)$
and $\myvec b_{\hat{a}}\triangleq-\eta_{v}\myvec h(t)\alpha$, with
$\alpha\in[0,1]$, such that
\begin{gather}
\left(\mymatrix B_{q}\dot{\myvec q}\preceq\bq\right)\land\left(\mymatrix B_{\hat{a}}\dot{\hat{\myvec a}}\preceq\bahat\right).\label{eq:separate-inequalities}
\end{gather}

\subsection{Closed-loop stability\label{subsec:Stability}}

To prove that the closed-loop system composed of \eqref{eq:estimated_dfkm},
\eqref{eq:measure_space_estimated_dfkm}, \eqref{eq:problem_quadratic_constrained},
and \eqref{eq:adaptive-constrained-law} is stable, we first define
\emph{equivalent} optimization problems to \eqref{eq:problem_quadratic_constrained}
and \eqref{eq:adaptive-constrained-law}. Then, we show that under
Assumptions 2 and 3 of Section~\ref{subsec:Problem-statement-and-assumptions},
those optimization problems are always feasible, which means that
it suffices to analyze the objective functions to determine closed-loop
stability. Finally, we choose a Lyapunov function candidate and exploit
the optimal solution to prove that the closed-loop system is stable.

\subsubsection*{Equivalent optimization problems}

We use the extended vector $\myvec v=\begin{bmatrix}\myvec q^{T} & \hat{\myvec a}^{T}\end{bmatrix}^{T}$
such that
\begin{gather}
\argminimone{\myvec u_{v,q}\in}{\dot{\myvec v}}{\norm{\mymatrix J_{\hat{x}}\dot{\myvec v}+\eta_{q}\estimatedtaskerror}_{2}^{2}+\norm{\myvec{\Lambda}_{v,q}\dot{\myvec v}}_{2}^{2}}{\begin{split}\myvec B_{v,q}\dot{\myvec v}\preceq\bq\\
\mymatrix W_{v,q}\dot{\myvec v}\preceq\myvec w_{q}\\
\mymatrix A_{q}\dot{\myvec v}=\myvec 0,
\end{split}
}\label{eq:equivalent task-space control law}
\end{gather}
in which $\mymatrix J_{\hat{x}}=\begin{bmatrix}\mymatrix J_{\hat{x},q} & \mymatrix J_{\hat{x},\hat{a}}\end{bmatrix}$,
$\myvec{\Lambda}_{v,q}=\begin{bmatrix}\mymatrix{\Lambda}_{q} & \mymatrix 0_{n\times p}\end{bmatrix}\in\mathbb{R}^{n\times(n+p)}$,
$\mymatrix B_{v,q}=\begin{bmatrix}\mymatrix B_{q} & \mymatrix 0_{s\times p}\end{bmatrix}\in\mathbb{R}^{s\times(n+p)}$,
$\mymatrix W_{v,q}=\begin{bmatrix}\mymatrix W_{q} & \mymatrix 0_{s_{q}\times p}\end{bmatrix}\in\mathbb{R}^{s_{q}\times(n+p)}$,
and $\mymatrix A_{q}=\mathrm{blkdiag}(\mymatrix 0_{n\times n},\mymatrix I_{p\times p})\in\mathbb{R}^{(n+p)\times(n+p)}$.
Notice that \eqref{eq:equivalent task-space control law} boils down
to \eqref{eq:problem_quadratic_constrained}. More specifically, $\mymatrix B_{v,q}\dot{\myvec v}=\mymatrix B_{q}\dot{\myvec q}$
and $\mymatrix W_{v,q}\dot{\myvec v}=\mymatrix W_{q}\dot{\myvec q}$.
Also, $\mymatrix A_{q}\dot{\myvec v}=\myvec 0$ implies $\dot{\hat{\myvec a}}=\myvec 0$.
(It also implies $\mymatrix 0_{n\times n}\dot{\myvec q}=\myvec 0$,
which is always true regardless the value of $\dot{\myvec q}$). Hence,
\begin{align*}
\mymatrix J_{\hat{x}}\dot{\myvec v}+\eta_{q}\estimatedtaskerror & =\mymatrix J_{\hat{x},q}\dot{\myvec q}+\mymatrix J_{\hat{x},\hat{a}}\dot{\hat{\myvec a}}+\eta_{q}\estimatedtaskerror=\mymatrix J_{\hat{x},q}\dot{\myvec q}+\eta_{q}\estimatedtaskerror.
\end{align*}
Lastly, $\myvec{\Lambda}_{v,q}\dot{\myvec v}=\myvec{\Lambda}_{q}\dot{\myvec q}$
and $\myvec u_{v,q}=\begin{bmatrix}\myvec u_{q}^{T} & \myvec 0^{T}\end{bmatrix}^{T}$.

Analogously,
\begin{gather}
\argminimone{\myvec u_{v,\estimated a}\in}{\dot{\myvec v}}{\norm{\mymatrix J_{\estimated y}\dot{\myvec v}+\eta_{\estimated a}\tilde{\myvec y}}_{2}^{2}+\norm{\mymatrix{\Lambda}_{v,\hat{a}}\dot{\myvec v}}_{2}^{2}}{\begin{split}\mymatrix B_{v,\hat{a}}\dot{\myvec v} & \preceq\bahat\\
\mymatrix W_{v,\estimated a}\dot{\myvec v} & \preceq\myvec w_{\estimated a}\\
\mymatrix N_{v,\hat{a}}\dot{\myvec v} & =\myvec 0\\
\estimatedtaskerror^{T}\mymatrix J_{\hat{x}}\dot{\myvec v} & \leq0\\
\mymatrix A_{\hat{a}}\dot{\myvec v} & =\myvec 0,
\end{split}
}\label{eq:equivalent-adaptive-constrained-law}
\end{gather}
in which $\mymatrix J_{\hat{y}}=\begin{bmatrix}\mymatrix J_{\hat{y},q} & \mymatrix J_{\hat{y},\hat{a}}\end{bmatrix}$,
$\mymatrix{\Lambda}_{v,\hat{a}}=\begin{bmatrix}\mymatrix 0_{p\times n} & \mymatrix{\Lambda}_{\hat{a}}\end{bmatrix}\in\mathbb{R}^{p\times(n+p)}$,
$\mymatrix B_{v,\hat{a}}=\begin{bmatrix}\mymatrix 0_{s\times n} & \mymatrix B_{\hat{a}}\end{bmatrix}\in\mathbb{R}^{s\times(n+p)}$,
$\mymatrix W_{v,\estimated a}=\begin{bmatrix}\mymatrix 0_{s_{\estimated a}\times n} & \mymatrix W_{\estimated a}\end{bmatrix}\in\mathbb{R}^{s_{\estimated a}\times(n+p)}$,
$\mymatrix N_{v,\hat{a}}=\begin{bmatrix}\mymatrix 0_{\bar{r}\times n} & \mymatrix N_{\hat{a}}\end{bmatrix}\in\mathbb{R}^{\bar{r}\times(n+p)}$,
and $\mymatrix A_{\hat{a}}=\mathrm{blkdiag}(\mymatrix I_{n\times n},\mymatrix 0_{p\times p})\in\mathbb{R}^{(n+p)\times(n+p)}$.
Again, it is easy to show that \eqref{eq:equivalent-adaptive-constrained-law}
is equivalent to \eqref{eq:adaptive-constrained-law}. First, by direct
calculation we find that the first three constraints in \eqref{eq:equivalent-adaptive-constrained-law}
boil down to the first three constraints in \eqref{eq:adaptive-constrained-law}.
Also, the constraint $\mymatrix A_{\hat{a}}\dot{\myvec v}=\myvec 0$
ensures that $\dot{\myvec q}=\myvec 0$. Therefore, $\estimatedtaskerror^{T}\mymatrix J_{\hat{x}}\dot{\myvec v}=\estimatedtaskerror^{T}\mymatrix J_{\hat{x},\estimated a}\dot{\hat{\myvec a}}$.
Lastly, because $\dot{\myvec q}=\myvec 0$, the objective function
in \eqref{eq:equivalent-adaptive-constrained-law} reduces to the
one in \eqref{eq:adaptive-constrained-law}, and $\myvec u_{v,\estimated a}=\begin{bmatrix}\myvec 0^{T} & \myvec u_{\hat{a}}^{T}\end{bmatrix}^{T}$.

\subsubsection*{Feasibility of the optimization problems}

Assumption~2 in Section~\ref{subsec:Problem-statement-and-assumptions}
determines that $\myvec v(0)\in\mathscr{F}$, which implies that $\myvec h(0)\preceq\myvec 0$
and, consequently, $\bq(0),\myvec b_{\hat{a}}(0)\succeq\myvec 0$.
Because $\dot{\myvec v}=\myvec 0$ is a feasible solution by Assumption~3,
the optimization problems \eqref{eq:equivalent task-space control law}
and \eqref{eq:equivalent-adaptive-constrained-law} are always feasible.
In other words, in the worst case, the robot can stop. Hence, it is
sufficient to analyze only the objective function to determine closed-loop
stability.
\begin{rem}
Gronwall's lemma \cite{gronwall} guarantees that the system composed
of \eqref{eq:estimated_dfkm}, \eqref{eq:measure_space_estimated_dfkm},
\eqref{eq:problem_quadratic_constrained}, and \eqref{eq:adaptive-constrained-law}
will never violate a constraint in the estimated task-space. However,
this does not mean that it is guaranteed that there will be no violation
in the real task-space. A practical solution to this problem is shown
in Section~\ref{sec:E2_validation}, which consists of giving the
system some time to adjust the estimated model to be closer to the
real one before the robot starts moving.
\end{rem}

\subsubsection*{Lyapunov stability\label{subsec:Lyapunov-stability}}

Choosing the Lyapunov candidate $V(\estimatedtaskerror)\triangleq V=\left(1/2\right)\estimatedtaskerror^{T}\estimatedtaskerror$,
in which $\estimatedtaskerror=\hat{\myvec x}-\myvec x_{d}$, we obtain
$\dot{V}=\estimatedtaskerror^{T}\dot{\hat{\myvec x}}=\estimatedtaskerror^{T}\mymatrix J_{\hat{x}}\dot{\myvec v}$.
Since $\myvec u_{v,q}=\begin{bmatrix}\myvec u_{q}^{T} & \myvec 0^{T}\end{bmatrix}^{T}$
and $\myvec u_{v,\estimated a}=\begin{bmatrix}\myvec 0^{T} & \myvec u_{\hat{a}}^{T}\end{bmatrix}^{T}$,
then $\dot{\myvec v}=\myvec u_{v,q}+\myvec u_{v,\estimated a}$. Therefore,
\begin{align}
\dot{V}=\estimatedtaskerror^{T}\mymatrix J_{\hat{x}}\dot{\myvec v} & =\estimatedtaskerror^{T}\mymatrix J_{\hat{x}}\myvec u_{v,q}+\estimatedtaskerror^{T}\mymatrix J_{\hat{x}}\myvec u_{v,\estimated a}.\label{eq:term that must be non positive}
\end{align}
Now we show that each term of the right-hand side of \eqref{eq:term that must be non positive}
is non-positive. From \eqref{eq:equivalent task-space control law},
we have
\begin{align*}
\norm{\mymatrix J_{\hat{x}}\myvec u_{v,q}+\eta_{q}\estimatedtaskerror}_{2}^{2}+\norm{\myvec{\Lambda}_{v,q}\myvec u_{v,q}}_{2}^{2} & \leq\norm{\eta_{q}\estimatedtaskerror}_{2}^{2}
\end{align*}
because $\myvec u_{v,q}$ is the optimal solution and $\dot{\myvec v}=\myvec 0$
is always a feasible solution by assumption. Therefore,\small 
\begin{align*}
\norm{\mymatrix J_{\hat{x}}\myvec u_{v,q}}_{2}^{2}+2\eta_{q}\estimatedtaskerror^{T}\mymatrix J_{\hat{x}}\myvec u_{v,q}+\norm{\eta_{q}\estimatedtaskerror}_{2}^{2}+\norm{\myvec{\Lambda}_{v,q}\myvec u_{v,q}}_{2}^{2} & \leq\norm{\eta_{q}\estimatedtaskerror}_{2}^{2},
\end{align*}
\normalsize which implies 
\begin{multline}
\estimatedtaskerror^{T}\mymatrix J_{\hat{x}}\myvec u_{v,q}\leq-\frac{1}{2\eta_{q}}\left(\norm{\mymatrix J_{\hat{x}}\myvec u_{v,q}}_{2}^{2}+\norm{\myvec{\Lambda}_{v,q}\myvec u_{v,q}}_{2}^{2}\right)\leq0.\label{eq:task-space term non positive}
\end{multline}
From \eqref{eq:equivalent-adaptive-constrained-law} we have that
$\estimatedtaskerror^{T}\mymatrix J_{\hat{x}}\myvec u_{v,\estimated a}\leq0$
is enforced by the optimization problem. Hence, \eqref{eq:term that must be non positive}
is nonpositive, which implies $\dot{V}=\estimatedtaskerror^{T}\mymatrix J_{\hat{x}}\dot{\myvec v}\leq0$.
Thus, we conclude that the closed-loop system is stable.
\begin{rem}
Due to the generality of the constraints being considered, only closed-loop
stability can be ensured. One intuitive way of illustrating this is
to think of a manipulator robot holding an object trying to place
it on a table nearby. If there is a wall between the robot and the
table, the constraint will prevent the robot from reaching the table,
making it physically impossible to have asymptotic convergence without
piercing through the wall. Notice that this is the desired behavior:
the safety of the robot and its environment are prioritized over asymptotic
convergence.
\end{rem}
\begin{rem}
When using non-additive errors (e.g., when using unit quaternions
to represent orientations or unit dual quaternions to represent poses),
it is possible to use suitable projectors such that the argument above
still holds. See Appendix~\ref{sec:Appendix-Stability}.
\end{rem}

\section{Use cases\label{sec:Use-cases}}

\begin{table}[h]
\caption{\textcolor{black}{\label{tab:Parametric-measure-space-estimat}Parametric
measure-space estimated Jacobian and $\protect\parametrictaskJcomplement$.}}

\begin{center}

\noindent\resizebox{0.8\columnwidth}{!}{%
\begin{centering}
\begin{tabular}{c||cccc}
 & $\measurespace_{x}$ & $\measurespace_{r}$ & $\measurespace_{t}$ & $\mathcal{Y}_{d}$\tabularnewline
\hline 
$\mymatrix J_{\estimated y,\hat{a}}$ & $\mymatrix J_{x,\hat{a}}$ & $\mymatrix J_{r,\hat{a}}$ & $\mymatrix J_{t,\hat{a}}$ & $\mymatrix J_{d,\hat{a}}$\tabularnewline
$\mymatrix N_{\hat{a}}$ & $\boldsymbol{0}$ & $\mymatrix J_{t,\hat{a}}$ & $\mymatrix J_{r,\hat{a}}$ & \eqref{eq:N_d}\tabularnewline
\end{tabular}
\par\end{centering}
}

\end{center}

\medskip{}

\textcolor{black}{Notice that $\mymatrix J_{x,\hat{a}}\in\mathbb{R}^{m_{x}\times n}$,
$\mymatrix J_{r,\hat{a}}\in\mathbb{R}^{m_{r}\times n}$, $\mymatrix J_{t,\hat{a}}\in\mathbb{R}^{3\times n}$,
and $\mymatrix J_{d,\hat{a}}\in\mathbb{R}^{1\times n}$ are the parametric
Jacobians that satisfy \eqref{eq:estimated_dfkm} for the estimated
pose, rotation, translation, and distance, respectively. The dimensions
$m_{x}$ and $m_{r}$ depend on the parameterization used for the
pose and rotation.}
\end{table}

We summarize four relevant use-cases for the measurement space $\measurespace$
(i.e., the space of relevant end-effector measurements with respect
to the sensor's reference frame $\frame w$), the estimated measure-space
Jacobian $\mymatrix J_{\estimated y,\hat{a}}$, and the $\parametrictaskJcomplement$
$\mymatrix N_{\hat{a}}$ in Table~\ref{tab:Parametric-measure-space-estimat}.
A more detailed description of how each measurement space is parameterized
is given in Appendix~\ref{sec:Implementation}.

\subsubsection*{Measurement of complete end-effector pose}

The measurement space is the end-effector pose space $\measurespace_{x}$
(Appendix~\ref{subsec:Poses-and-rigid}). Given that this represents
a complete measurement, the $\parametrictaskJcomplement$ is trivial,
i.e. $\mymatrix N_{\hat{a}}\triangleq\mymatrix 0$.

\subsubsection*{Measurement of end-effector orientation}

The measurement space is the end-effector rotation space $\measurespace_{r}$
(Appendix~\ref{subsec:Orientations-and-rotations}). In this case,
the real translation $\myvec t$ is unknown, so we enforce $\dot{\hat{\myvec t}}=\myvec 0$,
which is achieved when $\mymatrix N_{\hat{a}}\triangleq\mymatrix J_{t,\hat{a}}$.

\subsubsection*{Measurement of end-effector translation}

The measurement space is the end-effector translation space $\measurespace_{t}$
(Appendix~\ref{subsec:Positions-and-translations}). In this case,
the real rotation $\myvec r$ is unknown; hence, we constrain any
parameter update on the rotation by enforcing $\dot{\hat{\myvec r}}=\myvec 0$,
which is achieved when $\mymatrix N_{\hat{a}}\triangleq\mymatrix J_{r,\hat{a}}$.

\subsubsection*{Measurement of the end-effector distance}

\begin{figure}[t]
\centering

\def\svgwidth{0.4\columnwidth}

\small
\begingroup%
  \makeatletter%
  \providecommand\color[2][]{%
    \errmessage{(Inkscape) Color is used for the text in Inkscape, but the package 'color.sty' is not loaded}%
    \renewcommand\color[2][]{}%
  }%
  \providecommand\transparent[1]{%
    \errmessage{(Inkscape) Transparency is used (non-zero) for the text in Inkscape, but the package 'transparent.sty' is not loaded}%
    \renewcommand\transparent[1]{}%
  }%
  \providecommand\rotatebox[2]{#2}%
  \newcommand*\fsize{\dimexpr\f@size pt\relax}%
  \newcommand*\lineheight[1]{\fontsize{\fsize}{#1\fsize}\selectfont}%
  \ifx\svgwidth\undefined%
    \setlength{\unitlength}{165.90341643bp}%
    \ifx\svgscale\undefined%
      \relax%
    \else%
      \setlength{\unitlength}{\unitlength * \real{\svgscale}}%
    \fi%
  \else%
    \setlength{\unitlength}{\svgwidth}%
  \fi%
  \global\let\svgwidth\undefined%
  \global\let\svgscale\undefined%
  \makeatother%
  \begin{picture}(1,1.01772467)%
    \lineheight{1}%
    \setlength\tabcolsep{0pt}%
    \put(0,0){\includegraphics[width=\unitlength,page=1]{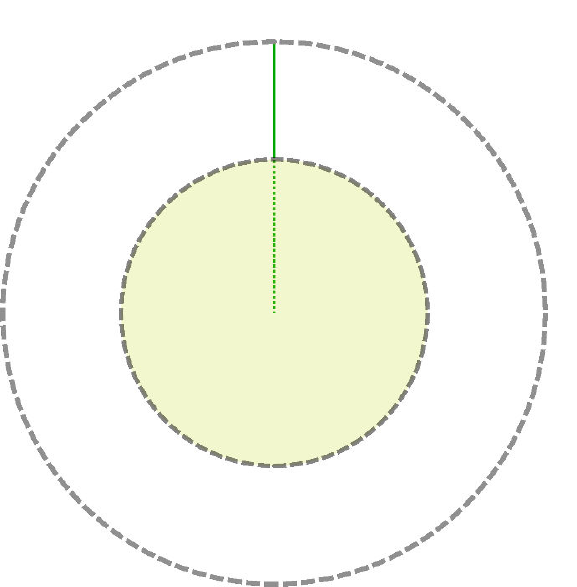}}%
    \put(0.4602705,0.97597881){\makebox(0,0)[lt]{\lineheight{1.25}\smash{\begin{tabular}[t]{l}$\hat{\myvec{t}}$\end{tabular}}}}%
    \put(0.22936176,0.68442734){\makebox(0,0)[lt]{\lineheight{1.25}\smash{\begin{tabular}[t]{l}$\myvec{t}$\end{tabular}}}}%
    \put(0,0){\includegraphics[width=\unitlength,page=2]{distance_invariant.pdf}}%
    \put(0.15719874,0.42594163){\makebox(0,0)[lt]{\lineheight{1.25}\smash{\begin{tabular}[t]{l}$d=R$\end{tabular}}}}%
    \put(0.83462028,0.66832319){\makebox(0,0)[lt]{\lineheight{1.25}\smash{\begin{tabular}[t]{l}$\hat{d}$\end{tabular}}}}%
    \put(0,0){\includegraphics[width=\unitlength,page=3]{distance_invariant.pdf}}%
    \put(0.50875747,0.76322734){\makebox(0,0)[lt]{\lineheight{1.25}\smash{\begin{tabular}[t]{l}$\myvec{t}_n$\end{tabular}}}}%
    \put(0,0){\includegraphics[width=\unitlength,page=4]{distance_invariant.pdf}}%
  \end{picture}%
\endgroup%

\caption{\label{fig:distance-invariant}All distances are measured with respect
to the origin of the sphere, and $\protect\myvec t,\hat{\protect\myvec t}$
$\in\mathbb{R}^{3}$ are the real and estimated positions, respectively.
The figure shows the circle formed by the intersection of the sphere
and the plane containing $\protect\myvec t$, $\hat{\protect\myvec t}$,
and the sphere's center. The solid green line represents all $\protect\myvec t_{\lambda}\in\mathbb{R}^{3}$
along the line that connects $\protect\estimated{\protect\myvec t}$
to the center, such that $R\protect\leq\protect\norm{\protect\myvec t_{\lambda}}_{2}\protect\leq\protect\norm{\hat{\protect\myvec t}}_{2}$
and $\hat{\protect\myvec t}\times\protect\myvec t_{\lambda}=\protect\myvec 0$.
Distance estimation updates in that line guarantee that the distance
estimation improves and the position estimation given by $\protect\norm{\hat{\protect\myvec t}-\protect\myvec t}_{2}$
does not worsen. The green point $\protect\myvec t_{n}$ satisfies
$\protect\norm{\protect\myvec t_{n}}=R$ with $\hat{\protect\myvec t}\times\protect\myvec t_{n}=\protect\myvec 0$.}
\end{figure}

The measurement space consists of the (Euclidean) distance space,
$\mathcal{Y}_{d}\triangleq\left\{ \norm{\quat p}_{2}:\quat p\in\mathbb{R}^{3}\right\} $,
which contains all possible distances between the end-effector position
and the inertial reference-frame. When we measure only the Euclidean
distance of the end-effector to the origin of the reference frame,
we constrain any rotation estimation update. In addition, we constrain
the position estimation updates to the line that connects the current
end-effector position to the origin of the reference frame, as shown
in Fig.~\ref{fig:distance-invariant}. This is because moving along
that line ensures that the estimated end-effector distance improves
while the position estimation does not worsen, as shown in the following
lemma. 
\begin{lem}
\label{lem:distance}Consider a sphere of radius $R$, an arbitrary
point $\myvec t$ on the sphere's surface, and a point $\estimated{\myvec t}$
outside of it. Also, let $\myvec t_{n}=R\estimated{\myvec t}/\norm{\estimated{\myvec t}}$
and $\lambda\in[0,1]$. Assuming the center of the sphere as the reference
point, the distance between any point $\myvec t_{\lambda}=(1-\lambda)\estimated{\myvec t}+\lambda\myvec t_{n}$
and $\myvec t$ is smaller or equal than the distance between $\estimated{\myvec t}$
and $\myvec t$.
\end{lem}
\begin{IEEEproof}
See Appendix~\ref{sec:Additional-proofs}.
\end{IEEEproof}
Therefore, when preventing the update of the rotation estimation and
constraining the position estimation update to the line that connects
the current end-effector position to the origin of the reference frame,
we obtain
\begin{align}
\begin{cases}
\dot{\hat{\myvec r}}=\myvec 0\\
\hat{\myvec t}\times\dot{\hat{\myvec t}}=\myvec 0
\end{cases} & \implies\underbrace{\begin{bmatrix}\mymatrix J_{r,\hat{a}}\\
\mymatrix S\left(\hat{\myvec t}\right)\mymatrix J_{t,\hat{a}}
\end{bmatrix}}_{\mymatrix N_{d,\hat{a}}}\dot{\hat{\myvec a}}=\myvec 0,\label{eq:N_d}
\end{align}
in which $\mymatrix S(\cdot):\mathbb{R}^{3}\to\mathrm{so}(3)$ is
a skew-symmetric matrix such that $\mymatrix S(\myvec a)\myvec b=\myvec a\times\mymatrix b$,
with $\myvec a,\myvec b\in\mathbb{R}^{3}$, and $\myvec t\in\mathbb{R}^{3}$
is the end-effector position with respect to $\frame w$. 

\section{Experiment PM: Partial Measurements\label{sec:E1_validation}}

\begin{table}[t]
\textcolor{black}{\caption{\label{tab:DH-parameters-of}DH parameters of the VS050 robot.}
}

\begin{center}

\noindent\resizebox{1.0\columnwidth}{!}{%

\begin{tabular}{ccccccc}
\hline 
 & 1 & 2 & 3 & 4 & 5 & 6\tabularnewline
\hline 
\hline 
$\theta_{\text{DH}}$ {[}rad{]} & $-\pi$ & $\pi/2$ & $-\pi/2$ & $0$ & $\pi$ & $0$\tabularnewline
$d_{\text{DH}}$ {[}m{]} & $0.345$ & $0$ & $0$ & $0.255$ & $0$ & $0.07$\tabularnewline
$a_{\text{DH}}$ {[}m{]} & $0$ & $0.250$ & $0.01$ & $0$ & $0$ & $0$\tabularnewline
$\alpha_{\text{DH}}$ {[}rad{]} & $\pi/2$ & 0 & $-\pi/2$ & $\pi/2$ & $\pi/2$ & 0\tabularnewline
\hline 
\end{tabular}

}

\end{center}
\end{table}

\begin{figure}[t]
\centering

\includegraphics[width=1\columnwidth]{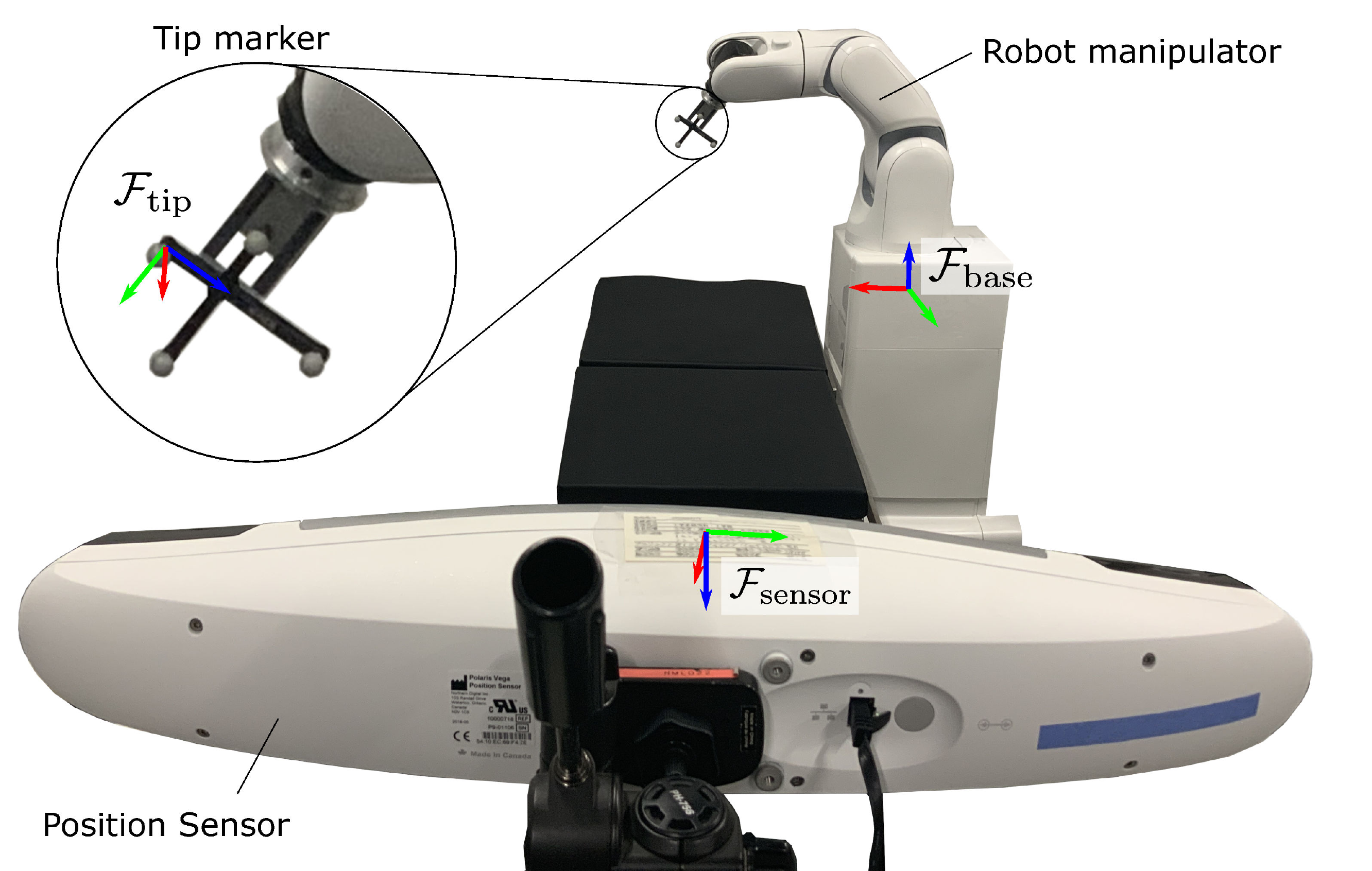}

\caption{\label{fig:setup} Visualization of the experimental setup.}
\end{figure}

The experimental setup\footnote{Software implementation based on \url{https://github.com/SmartArmStack}
and DQ Robotics C++11 \cite{adorno2020dqrobotics}.} is composed of a six-DoF manipulator robot (VS050, DENSO WAVE, Japan)
with coordinate system $\frame{\text{base}}$ and an image-based position
sensor (Polaris Vega, NDI, Canada) with coordinate system $\frame{\text{sensor}}$.
The image-based sensor was used to obtain the pose of the \emph{tip
marker} with respect to $\frame{\text{sensor}}$. The tip marker was
attached to the tip of the robotic manipulator and is represented
by $\frame{\text{tip}}$. These elements are shown in Fig.~\ref{fig:setup}.

In this first experiment, the objective is to reach four consecutive
setpoints in the robot's real task-space, which consists of the space
of end-effector poses (i.e., position and orientation), while adapting
the parameters of the robot. There is no collision avoidance because
the purpose is to evaluate the closed-loop system behavior under uncertain
parameters and complete or partial measurements. The parameters of
the robot consist of all the robot's DH parameters and an initial
rough estimate of $\frame{\text{base}}$ and $\frame{\text{tip}}$.

We used the manufacturer's documentation,\footnote{\url{https://www.denso-wave.com/en/robot/product/five-six/vs050-060.html}}
summarized in Table~\ref{tab:DH-parameters-of}, to obtain the initial
value for the DH parameters.\footnote{We chose the DH parameters because they are ubiquitous in the literature
on robot manipulators, but our methodology does not depend on specific
parameterization.} The control law adapts the parameters $\theta_{\text{DH}},d_{\text{DH}},a_{\text{DH}},$
and $\alpha_{\text{DH}}$ of each joint, resulting in 24 joint-related
parameters. The limits for the estimated translational parameters
are of $\pm1$~mm, and $\pm1^{\circ}$ for the angular parameters.

The transformations $\frame{\text{base}}$ and $\frame{\text{tip}}$
are modeled using six parameters each. Those parameters describe six
sequential transformations, namely a translation along the $x$-axis,
a translation along the $y$-axis, a translation along the $z$-axis,
a rotation about the $x$-axis, a rotation about the $y$-axis, and
a rotation about the $z$-axis. These two transformations were roughly
measured using a tape measure. The limits for the estimate of the
translation parameters are $\pm10$~cm from the initially measured
values, and the limits for the estimate of the angular parameters
are $\pm20^{\circ}$ from the initially measured values. With these
12 parameters in conjunction with the joint-related parameters, a
total of 36 parameters were estimated online (i.e., $p=36$).

The experiments were executed under five conditions. Without any adaptation
(\textbf{PM0}), and with adaptation using four subsets of the pose
measurement: (\textbf{PM1}) pose, (\textbf{PM2}) rotation, (\textbf{PM3})
translation, and (\textbf{PM4}) distance. Although in \textbf{PM2–4}
only partial measurements are used in the adaptation law, we store
the information of the complete end-effector pose measurements and
use them as ground-truth in our analyses.

The control loop runs at $50$~Hz ($T=20\,\unit{ms}$), which is
the maximum sampling frequency of our measurement system. The control
gains are $\eta_{q}=\eta_{\estimated a}=40$, and the damping factors
are $\mymatrix{\Lambda}_{q}=0.01\mymatrix I_{6}$ and $\mymatrix{\Lambda}_{\hat{a}}=0.01\mymatrix I_{36}$.
The joint velocity limits were set at $\pm0.2$~rad/s to partially
compensate for the relatively low sampling time.

\subsection*{Results and discussion}

\begin{figure*}[tbh]
\begin{centering}
\includegraphics[width=1\textwidth]{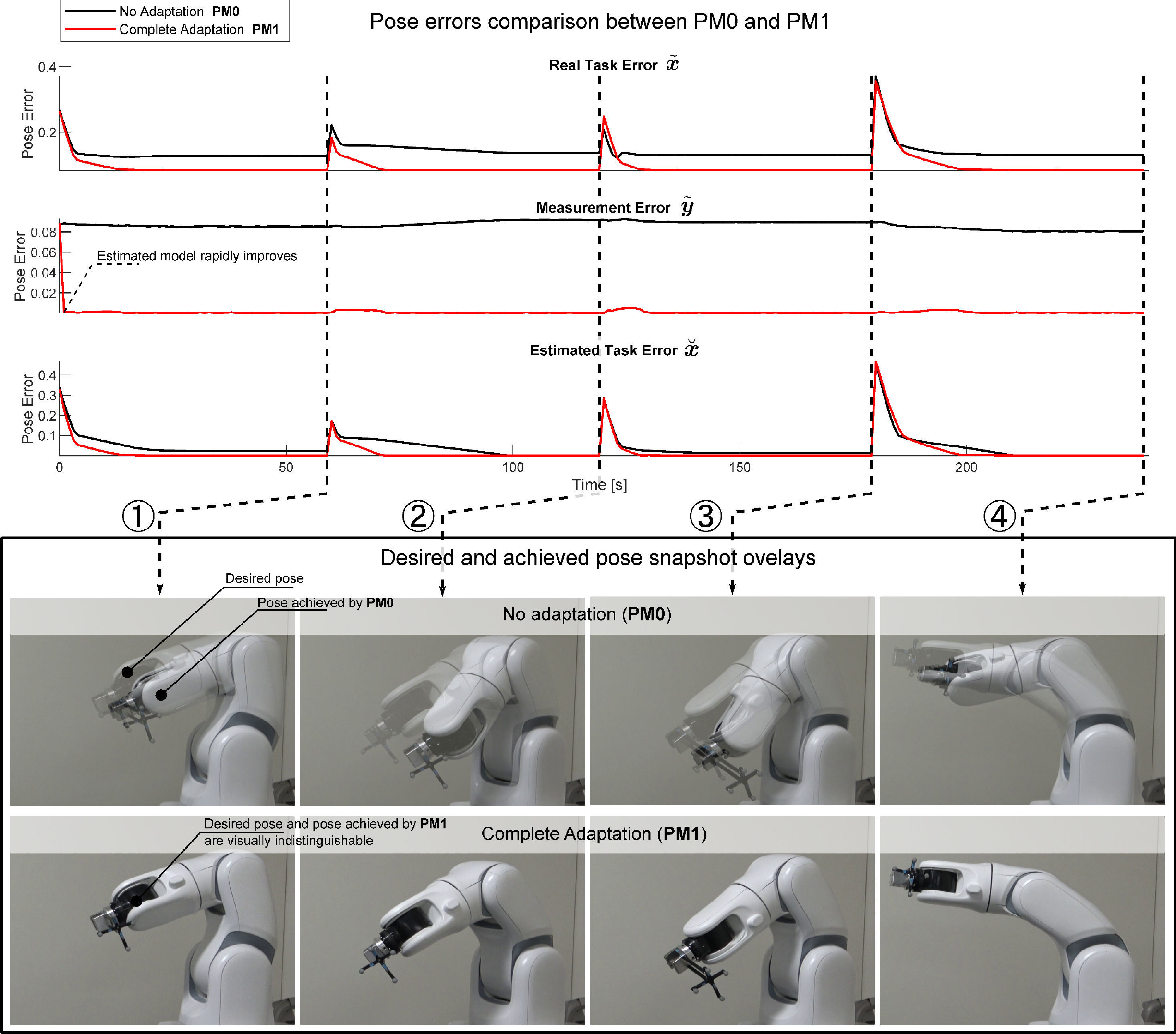}
\par\end{centering}
\caption{\label{fig:comparison_pm0_pm1}Comparison of the pose error between
\textbf{PM0} and \textbf{PM1}, in terms of real task pose error $\protect\realtaskerror$,
estimated task pose error $\protect\estimatedtaskerror$, and measurement
pose error $\protect\measurementerror$. The proposed controller with
complete measurements (\textbf{PM1}) outperforms a state-of-the-art
kinematic controller \cite{Marinho2019} without adaptation (\textbf{PM0})
in all errors. The snapshots of the experiment show qualitatively
the large final real task error in \textbf{PM0}, whereas the final
real task error in \textbf{PM1} is indistinguishable from the real
desired end-effector pose.}
\end{figure*}

The results regarding \textbf{PM0} and \textbf{PM1} are summarized
in Fig.~\ref{fig:comparison_pm0_pm1} in terms of real task error
$\realtaskerror$ (i.e., the error between the current measurement
$\myvec y$ and the desired pose $\myvec x_{d}$), the measurement
error $\measurementerror$ (i.e., the error between the current estimated
pose $\estimated{\myvec x}$ and the current measurement $\myvec y$),
and the estimated task error $\estimatedtaskerror$ (i.e., the error
between the current estimated pose $\estimated{\myvec x}$ and the
desired pose $\myvec x_{d}$). Notice that when the pose error norm
equals zero, the translation, rotation, and error norms also equal
zero.

\textcolor{blue}{}
\begin{figure*}[!p]
\begin{centering}
\textcolor{black}{\includegraphics[width=0.9\textwidth]{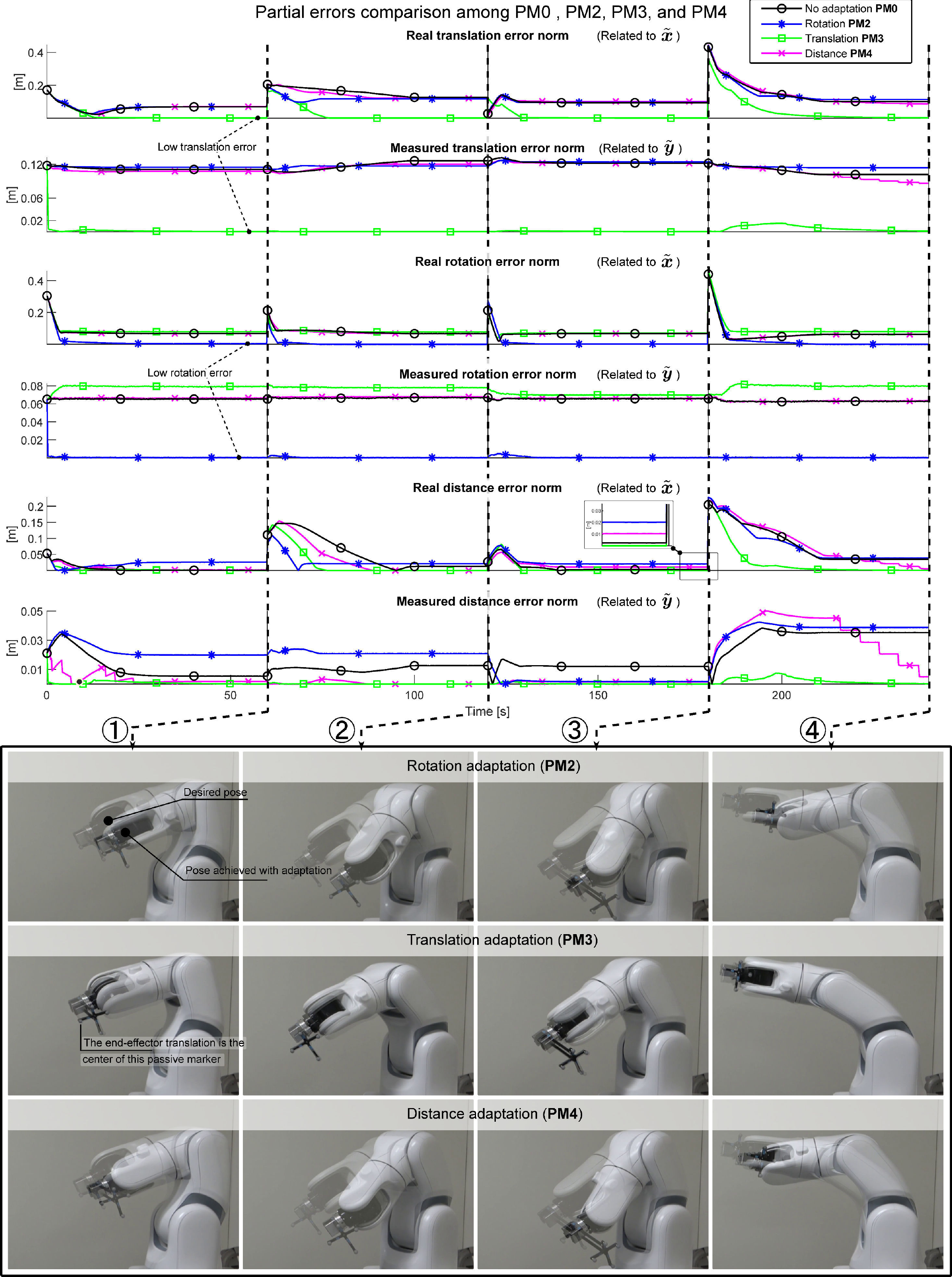}}
\par\end{centering}
\textcolor{black}{\caption{\label{fig:comparison_pm0_pm2_pm3_pm4}Comparison of the partial
errors among \textbf{PM0},\textbf{ PM2},\textbf{ PM3}, and \textbf{PM4},
in terms of real task pose error $\protect\realtaskerror$ and measurement
pose error $\protect\measurementerror$. The errors are defined in
Appendix~\ref{subsec:Error-definition}. The controllers with partial
measurements \textbf{PM2} and \textbf{PM3} achieve their real targets
in translation and rotation, respectively, for all setpoints. The
controller with partial measurements \textbf{PM4} has a distance error
of several millimeters in some setpoints given that the nominal model
was imprecise and the robot got stuck near a joint limit (see Fig.~\ref{fig:pm4_stuck}).}
}
\end{figure*}

Regarding partial measurements, Fig.~\ref{fig:comparison_pm0_pm2_pm3_pm4}
shows the results of \textbf{PM0},\textbf{ PM2},\textbf{ PM3}, and\textbf{
PM4} in terms of translation, rotation, and distance errors. Notice
that when the translation error norm equals zero, the distance error
norm also equals zero. The converse, however, is not true because
the distance error can be zero while the translation error is not
zero.

In our experiments, a given measurement space always caused the convergence
of the real task error and measurement error in that specific dimension
when the joint limits were not reached. For example, when the rotation
is measured, the real rotation error norm goes asymptotically to zero,
as shown by the solid-blue starred curve of the third graph in Fig.~\ref{fig:comparison_pm0_pm2_pm3_pm4},
and the measurement rotation error norm also goes to zero, as shown
by the solid-blue starred curve of the forth graph in Fig.~\ref{fig:comparison_pm0_pm2_pm3_pm4}.

\begin{figure}[tbh]
\begin{centering}
\includegraphics[width=1\columnwidth]{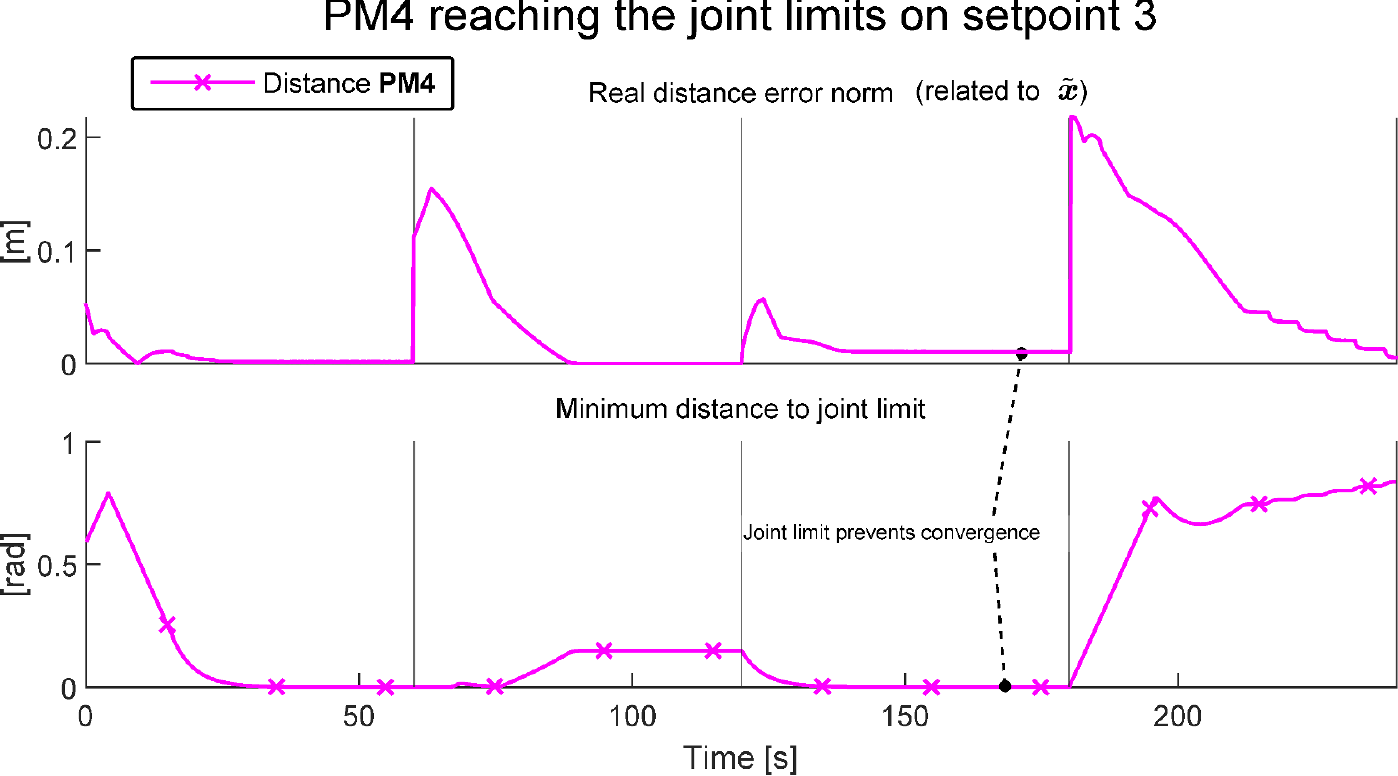}
\par\end{centering}
\caption{\label{fig:pm4_stuck} The plot of the real distance error for \textbf{PM4}
and the distance to a joint limit. Notice that in setpoint 3 the robot
is unable to converge in distance given the proximity to a joint limit.}
\end{figure}

\subsubsection*{No adaptation}

\textbf{PM0} represents the performance of a task controller without
any parameter adaptation, which means that only \eqref{eq:problem_quadratic_constrained}
was used in the generation of control inputs. The real task error
for all setpoints in Fig.~\ref{fig:comparison_pm0_pm1} illustrates
the errors in the initial estimation of the parameters. Because the
robot parameters are all initially incorrect, no adaptation means
that even if the estimated task error converges to zero, as shown
by the solid-black curves in the third graph of Fig.~\ref{fig:comparison_pm0_pm1}
during the regulation to the second and fourth setpoints, there is
no guarantee that the real-task error norm will converge to zero.
The estimated task error could not converge for setpoints 1 and 3
because the robot reached its joint limits. This happened in spite
of all setpoints being reachable in the real task-space. The measurement
error when there is no adaptation can be understood as the results
of a state-of-the-art constrained kinematic controller without adaptation
\cite{Marinho2019}, that we use as a comparative baseline for our
adaptive strategies.

\subsubsection*{Measurement of complete end-effector pose}

The complete pose measurement \textbf{(PM1}), represented by the solid
red circled curves in Fig.~\ref{fig:comparison_pm0_pm1}, resulted
in the convergence to zero of the real task error, the measurement
error, and the estimated task error for all setpoints. Because of
that, the translation, rotation, and error norms also converge to
zero.

\subsubsection*{Measurement of end-effector orientation}

When only rotation is measured (\textbf{PM2}), there was convergence
of the estimated rotation error. It caused a reduction of the real
pose error and the pose measurement error. The real and measurement
translation errors remained similar to the unmeasured cases, thanks
to the projection into the nullspace of the translation Jacobian,
but the real and measurement distance errors worsened (fifth and sixth
graphs in Fig.~\ref{fig:comparison_pm0_pm2_pm3_pm4}). Although this
may seem counterintuitive, it is not unexpected. Indeed, let us consider
the estimated position $\estimated{\myvec p}$ and real position $\myvec p$
associated with the estimated end-effector pose $\estimated{\myvec x}$
and real end-effector pose $\myvec x$, respectively. The translation
measurement error is given by $\tilde{\quat p}=\hat{\quat p}-\quat p$
and the distance measurement error is given by $\tilde{d}=\norm{\hat{\myvec p}}-\norm{\myvec p}$.
Hence, $\norm{\tilde{\myvec p}}^{2}-\norm{\tilde{d}}^{2}=-2\estimated{\myvec p}^{T}\myvec p+2\norm{\hat{\myvec p}}\norm{\myvec p}$.
Since $\estimated{\myvec p}^{T}\myvec p=\norm{\hat{\myvec p}}\norm{\myvec p}\cos\phi$,
where $\phi$ is the angle between $\hat{\myvec p}$ and $\myvec p$,
we obtain $\norm{\tilde{d}}^{2}=\norm{\tilde{\myvec p}}^{2}-2\norm{\hat{\myvec p}}\norm{\myvec p}\left(1-\cos\phi\right)$.
Now, consider another estimated position $\estimated{\myvec p}'$
and the associated translation measurement error $\tilde{\quat p}'=\hat{\quat p}'-\quat p$,
such that $\norm{\tilde{\myvec p}}=\norm{\tilde{\myvec p}'}$. Analogously,
let $\tilde{d}'=\norm{\hat{\myvec p}'}-\norm{\myvec p}$, then $\norm{\tilde{d}'}^{2}=\norm{\tilde{\myvec p}'}^{2}-2\norm{\hat{\myvec p}'}\norm{\myvec p}\left(1-\cos\phi'\right)$,
where $\phi'$ is the angle between $\hat{\myvec p}'$ and $\myvec p$.
Thus, 
\[
\norm{\tilde{d}'}^{2}=\norm{\tilde{d}}^{2}+2\norm{\myvec p}\left(\norm{\hat{\myvec p}}\left(1-\cos\phi\right)-\norm{\hat{\myvec p}'}\left(1-\cos\phi'\right)\right)
\]
 because $\norm{\tilde{\myvec p}}=\norm{\tilde{\myvec p}'}$. If $\norm{\hat{\myvec p}}\left(1-\cos\phi\right)>\norm{\hat{\myvec p}'}\left(1-\cos\phi'\right)$,
then $\norm{\tilde{d}'}^{2}>\norm{\tilde{d}}^{2}$, which implies
$\norm{\tilde{d}'}>\norm{\tilde{d}}$.

\subsubsection*{Measurement of end-effector translation}

When only the translation is measured (\textbf{PM3}), the real translation
error and the measured translation error, indicated by the green squared
curves in Fig.~\ref{fig:comparison_pm0_pm2_pm3_pm4} converge to
zero. Since zero translation error norm implies a zero distance error
norm, the distance error also converged to zero. A better estimation
of the translation allowed the estimated end-effector pose to converge
to all setpoints without reaching the joint limits.

\subsubsection*{Measurement of the end-effector distance}

In \textbf{PM4}, using the distance measurements only allowed for
the convergence of measured distance, but not the real and estimated
distances in setpoint 3 because the joint limits were reached, as
illustrated by the pink crossed curves in Figs.~\ref{fig:pm4_stuck}
and \ref{fig:comparison_pm0_pm2_pm3_pm4}.

These results correspond to the theoretical expectations. When the
robot was unable to reach specific setpoints, it gracefully stopped
moving in accordance with our proof of closed-loop stability. They
also show the benefits of using partial information when only that
information would be available. Depending on the requirements of a
given task, the designer should choose an appropriate sensor by considering
the trade-off between the expected improvement in real task error
and the monetary/time costs related to using that sensor.

\section{Experiment CA: Collision Avoidance\label{sec:E2_validation}}

\begin{algorithm}[tbh]
\small
\algnewcommand\algorithmicforeach{\textbf{for each}} 
\algdef{S}[FOR]{ForEach}[1]{\algorithmicforeach\ #1\ \algorithmicdo}
\algnewcommand{\LineComment}[1]{\State \(\triangleright\) #1}
\makeatletter \renewcommand{\ALG@name}{Algorithm} \makeatother
\begin{algorithmic}[1]

\State$T\gets$ sampling time

\State$\dq x_{\text{target}}\gets$ filtered target measurement 

\State$\dq x_{\text{inter}}\gets\mathrm{getIntermediateTarget}\left(\dq x_{\text{target}}\right)$

\State$\dq x_{\text{box}}\gets$ filtered box measurement

\State$\estimated{\myvec a}\gets$ initially estimated parameters

\While{not $\mathrm{isPlausible}\left(\myvec q,\estimated{\myvec a},\dq x_{\text{box}}\right)$}

\State$\estimated{\myvec a}\gets U\left(\estimated{\myvec a}_{\text{min}},\estimated{\myvec a}_{\text{max}}\right)$

\EndWhile

\For{$\dq x_{d}$ in $\left\{ \dq x_{\text{inter}},\dq x_{\text{target}}\right\} $}

\While{not $\mathrm{stopCriterion()}$}

\State$\myvec q\gets$ current robot configuration

\State$\estimatedtaskerror\gets\mathrm{getTaskError}\left(\myvec q,\estimated{\myvec a},\dq x_{d}\right)$

\State$\mymatrix J_{\hat{x},q}\gets\mathrm{getJacobianQ}\left(\myvec q,\estimated{\myvec a}\right)$

\State$\left(\myvec W_{q},\myvec w_{q}\right)\gets\mathrm{getLimitsQ}\left(\myvec q,\myvec q_{\text{min}},\myvec q_{\text{max}}\right)$

\State$\left(\myvec B_{q},\bq\right)\gets\mathrm{getVFIsQ}\left(\myvec q,\estimated{\myvec a},\dq x_{\text{box}}\right)$

\LineComment{For the control law, see \eqref{eq:problem_quadratic_constrained}}

\State $\myvec u_{q}\gets\mathrm{controlLaw}\left(\estimatedtaskerror,\mymatrix J_{\hat{x},q},\myvec B_{q},\bq,\myvec W_{q},\myvec w_{q}\right)$

\State$\myvec y\gets$ current task-space measurement

\If{$\myvec y$ is valid}

\State$\tilde{\myvec y}\gets\mathrm{getMeasurementError}\left(\myvec q,\estimated{\myvec a},\dq y\right)$

\State$\left(\mymatrix J_{\hat{x},\estimated a},\mymatrix J_{\estimated y,\estimated a}\right)\gets\mathrm{getJacobiansA}\left(\myvec q,\estimated{\myvec a}\right)$

\State$\mymatrix N_{\hat{a}}\gets\mathrm{getParametricJacobianProjector}\left(\myvec q,\estimated{\myvec a}\right)$

\State$\left(\myvec W_{\hat{a}},\myvec w_{\hat{a}}\right)\gets\mathrm{getLimitsA}\left(\hat{a},\hat{a}_{\text{min}},\hat{a}_{\text{max}}\right)$

\State$\left(\mymatrix B_{\hat{a}},\bahat\right)\gets\mathrm{getVFIsA}\left(\myvec q,\estimated{\myvec a},\dq x_{\text{box}}\right)$

\LineComment{For the adaptation law, see \eqref{eq:adaptive-constrained-law}}

\State $\begin{array}{c}
\myvec u_{\hat{a}}\gets\mathrm{adaptationLaw}(\estimatedtaskerror,\tilde{\myvec y},\mymatrix J_{\hat{x},\estimated a},\mymatrix J_{\estimated y,\estimated a}\\
\mymatrix B_{\hat{a}},\bahat,\myvec W_{\hat{a}},\myvec w_{\hat{a}})
\end{array}$

\State$\estimated{\myvec a}\leftarrow\estimated{\myvec a}+T\myvec u_{\hat{a}}$

\EndIf

\State$\myvec q\leftarrow\myvec q+T\quat u_{q}$

\State$\mathrm{sendToRobot}\left(\myvec q\right)$

\State$\mathrm{sleepUntilNextLoop}\left(\right)$

\EndWhile

\EndFor

\end{algorithmic}

\caption{\label{alg:experiment_CA} Proposed task-space constrained adaptive
control strategy implemented in Section~\ref{sec:E2_validation}.}
\end{algorithm}

\begin{figure}[t]
\begin{centering}
\includegraphics[width=1\columnwidth]{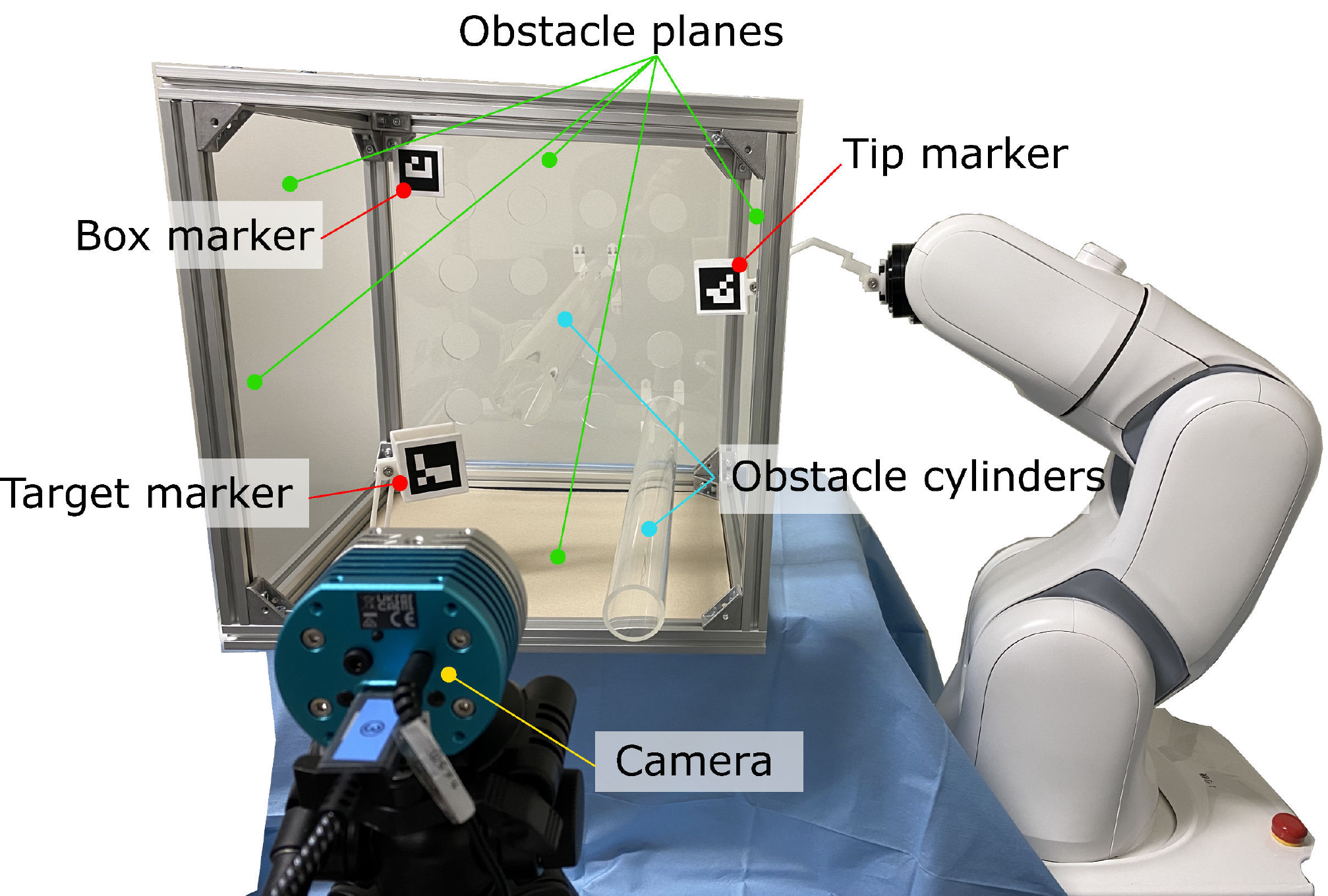}
\par\end{centering}
\caption{\textcolor{blue}{\label{fig:E2_setup}}Experimental setup for the
collision avoidance experiment \textbf{(CA}).}
\end{figure}

We conducted experiments to evaluate the behavior of the system when
there are obstacles in the workspace. The parameters and parameter
boundaries for the 24 joint-related parameters and six base-related
parameters were the same as the ones described in Section~\ref{sec:E1_validation}.
The end-effector for this experiment was a custom-designed intricate-shaped
3D-printed holder for ARUCO fiducial markers \cite{Garrido_Jurado_2014}.
The six end-effector-related parameters were obtained from the CAD
design and parameter boundaries were set as $\pm1$~cm for the translation-related
parameters and $\pm5^{\circ}$ for the rotation-related parameters.

The goal for the experiment was to move the end-effector, from the
initial pose, to a target pose $\dq x_{d,1}$ in the real task-space
while avoiding collisions with the obstacles, and then insert the
end-effector into a narrow slit. The target pose was obtained in real-time
from a custom-designed 3D-printed holder with an ARUCO marker. To
facilitate the insertion of the end-effector into the slit, we defined
an intermediate pose, $\dq x_{d,0}$, with the same rotation as $\dq x_{d,1}$,
but displaced $15$~cm. The obstacles were four planes and two four-centimeter-diameter
cylinders attached to a $40$~$\text{cm}^{3}$ cube made of aluminum
frames. The position of these obstacles were obtained with respect
to an ARUCO marker placed on a 3D-printed holder attached to the cube.
The experimental setup is shown in Fig.~\ref{fig:E2_setup}.

Measurements of the three ARUCO markers were obtained from a camera
(STC-HD853HDMI, Omron Sentech, Japan) and lens (VS-LDA4, Omron Sentech,
Japan) set up for 1080p 60~Hz readings using a PCI-E capture board
(Decklink Quad HDMI Recorder, Blackmagic Design, Australia). The camera
was calibrated using MATLAB's Camera Calibration application.\footnote{https://www.mathworks.com/help/vision/camera-calibration.html}
The ARUCO recognition was implemented using OpenCV\footnote{https://docs.opencv.org/4.x/d9/d53/aruco\_8hpp.html}
at 50~Hz. The end-effector measurement was obtained online without
filtering. The target pose and the cube pose were obtained using a
filter based on dual-quaternion (spatial) averaging \cite{Adorno2012}.

In the context of collision avoidance, the end-effector was enclosed
by six spheres. This is a conservative approach, but it satisfactorily
illustrates how we can use VFIs within the proposed adaptive formulation
to prevent collisions between the end-effector and the obstacles in
the workspace. The obstacles were two cylinders and four planes. To
prevent collisions between the end-effector spheres and the obstacle
cylinders, we used 12 point-to-line constraints. Those constraints
require the calculation of the squared distance $D_{t_{i},l_{j}}$
\cite[ Eq.  (29)]{Marinho2019} between the six end-effector's spheres
centered at $\quat t_{i}$, $i\in\left\{ 1,2,3,4,5,6\right\} $, and
the two obstacle cylinders' centerlines $\dq l_{j}$, $j\in\left\{ 1,2\right\} $,
in addition to the Jacobians $\mymatrix J_{t_{i},l_{j},q}$ and $\mymatrix J_{t_{i},l_{j},\estimated a}$
\cite[ Eq.  (32)]{Marinho2019} that satisfy $\mymatrix J_{t_{i},l_{j},q}\dot{\myvec q}+\mymatrix J_{t_{i},l_{j},\estimated a}\dot{\estimated{\myvec a}}=\dot{D}_{t_{i},l_{j}}$.
To prevent collisions between the end effector and the walls, we used
24 point-to-plane constraints using the distance $d_{t_{i},\pi_{k}}$
\cite[ Eq.  (57)]{Marinho2019} between the six end-effector spheres
and the four obstacle planes $\dq{\pi}_{k}$, $k\in\left\{ 1,2,3,4\right\} $,
in addition to the Jacobians $\mymatrix J_{t_{i},\pi_{k},q}$ and
$\mymatrix J_{t_{i},\pi_{k},\estimated a}$ \cite[ Eq.  (59)]{Marinho2019}
that satisfy $\mymatrix J_{t_{i},\pi_{k},q}\dot{\myvec q}+\mymatrix J_{t_{i},\pi_{k},\estimated a}\dot{\estimated{\myvec a}}=\dot{d}_{t_{i},\pi_{k}}$.
With these definitions, the VFI constraint in Problem~\ref{eq:problem_quadratic_constrained}
is composed of the following 36 inequalities
\[
\overbrace{\begin{bmatrix}-\mymatrix J_{t_{1},l_{1},q}\\
\vdots\\
-\mymatrix J_{t_{6},l_{2},q}\\
-\mymatrix J_{t_{1},\pi_{1},q}\\
\vdots\\
-\mymatrix J_{t_{6},\pi_{4},q}
\end{bmatrix}}^{\mymatrix B_{q}}\dot{\myvec q}\preceq\underbrace{\eta_{\text{vfi},q}\overbrace{\begin{bmatrix}D_{t_{1},l_{1}}-D_{\text{safe},t_{1},l}\\
\vdots\\
D_{t_{6},l_{2}}-D_{\text{safe},t_{6},l}\\
d_{t_{1},\pi_{1}}-d_{\text{safe},t_{1},\pi}\\
\vdots\\
d_{t_{6},\pi_{4}}-d_{\text{safe},t_{6},\pi}
\end{bmatrix}}^{\myvec h}}_{\bq}
\]
and the VFI constraint in Problem~\ref{eq:adaptive-constrained-law}
is composed of the following 36 inequalities
\[
\overbrace{\begin{bmatrix}-\mymatrix J_{t_{1},l_{1},\estimated a}\\
\vdots\\
-\mymatrix J_{t_{6},l_{2},\estimated a}\\
-\mymatrix J_{t_{1},\pi_{1},\estimated a}\\
\vdots\\
-\mymatrix J_{t_{6},\pi_{4},\estimated a}
\end{bmatrix}}^{\mymatrix B_{\estimated a}}\dot{\estimated{\myvec a}}\preceq\overbrace{\eta_{\text{vfi},\estimated a}\myvec h}^{\bahat},
\]
where $D_{\mathrm{safe},t_{i},l}=\left(R_{t_{i}}+R_{l}\right)^{2}$,
with $R_{t_{i}}\in\left\{ 0.04,0.015,0.015,0.015,0.015,0.075\right\} $~m
and $R_{l}=0.02$~m, and $d_{\mathrm{safe},t_{i},\pi}=R_{t_{i}}+d_{\mathrm{\pi}}$,
with $d_{\pi}=0.02$. The control gains were $\eta_{q}=\eta_{\estimated a}=4$
and the damping factors were $\mymatrix{\Lambda}_{q}=0.01\mymatrix I_{6}$
and $\mymatrix{\Lambda}_{\hat{a}}=0.01\mymatrix I_{36}$. The VFI
gains were $\eta_{\text{vfi},q}=\eta_{\text{vfi},\estimated a}=10$.
The joint velocity limits were set at $\pm0.01$~rad/s to partially
compensate for the relatively low sampling time of the ARUCO measurements.

The initial estimated parameters were obtained by sampling from a
uniform distribution within the parameter bounds until we obtained
a set of parameters that did not indicate collisions with obstacles.
This is necessary because although the actual robot is not in collision
with the obstacles at $t=0\,\unit{s}$, the \emph{estimated} robot
might be due to the uncertainties in the nominal model. In addition,
the robot was given $10$~s to update the parameters without motion
before moving from the initial pose to $\dq x_{d,0}$. This initial
adaptation allows the estimated model to get reasonably close to the
real model before the robot starts moving, which simplifies the tuning
of the control and adaptation gains (notice that $\eta_{q}=\eta_{\estimated a}$
and $\eta_{\text{vfi},q}=\eta_{\text{vfi},\estimated a}$). After
that, the robot was moved in sequence to $\dq x_{d,0}$ and $\dq x_{d,1}$.
The controller was executed during 150~s before moving to the next
set-point. The implementation for the experiment is summarized in
Algorithm~\ref{alg:experiment_CA}.

\subsection*{Results and discussion}

\begin{figure*}[!p]
\begin{centering}
\textcolor{black}{\includegraphics[width=0.95\textwidth]{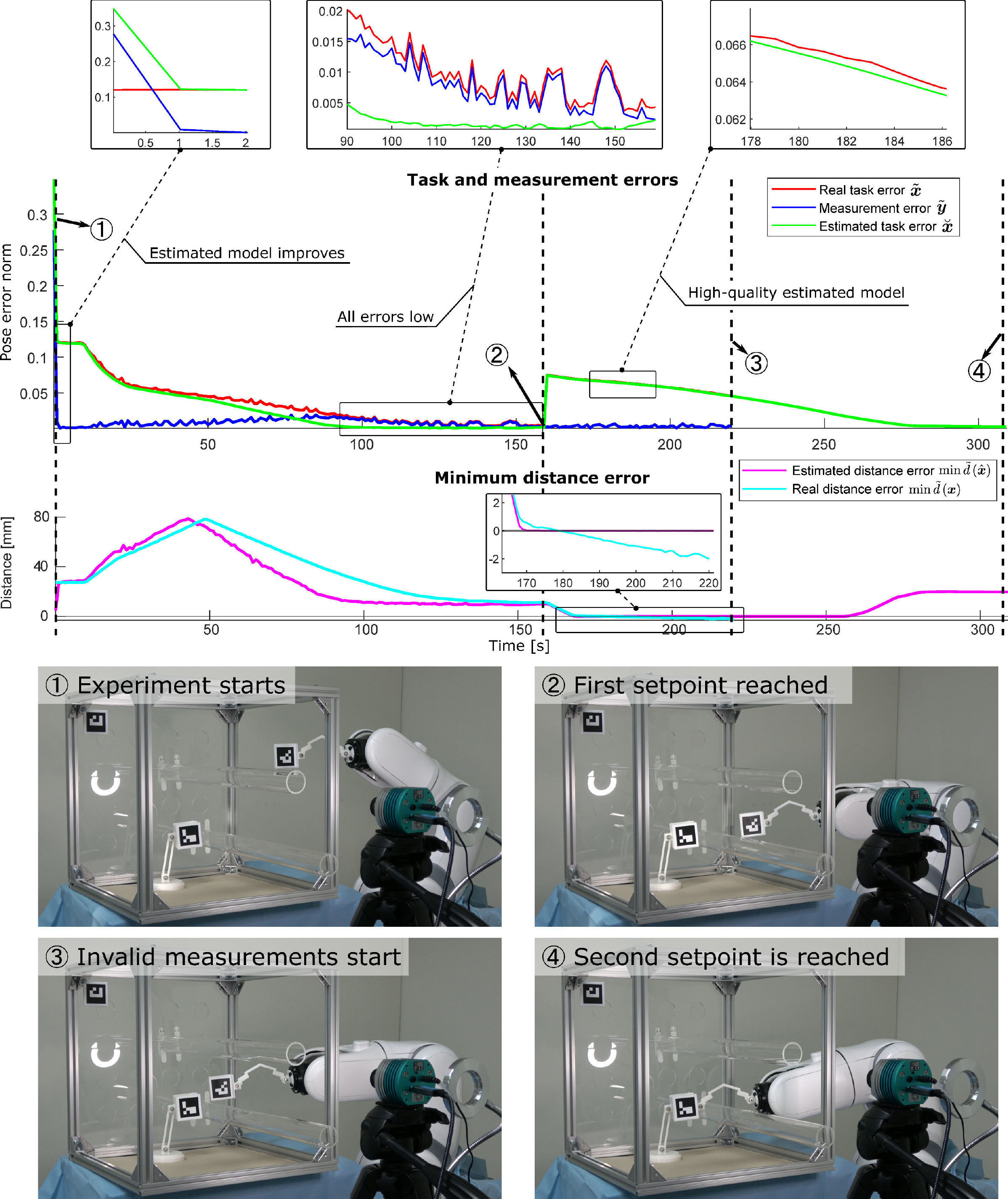}}
\par\end{centering}
\textcolor{black}{\caption{\label{fig:E2_results}\textcolor{black}{Results of the experiment
described in Section~\ref{sec:E2_validation} in terms of real task
error $\protect\realtaskerror$, estimated task error $\protect\estimatedtaskerror$,
measurement error $\protect\measurementerror$, and minimum distance
to obstacles. The minimum distance error is the minimum distance obtained
from all 36 collision pairs and a negative value indicates a penetration.
Snapshot \ding{172} stands for the beginning of the experiment, where
the end-effector was outside the box with a poorly estimated initial
model. For 10 seconds, the parameters were updated using the ARUCO
readings without moving the robot. This caused a near convergence
of the measurement error and sharply improved the estimated model.
The robot was then moved while adapting the parameters until snapshot
\ding{173}, where the robot converged to the first setpoint $\protect\dq x_{d,0}$.
At that point, all errors show convergent behavior and the setpoint
was changed. The robot used the high-quality model obtained so far
to move to the second setpoint, while continuously adapting the model
until snapshot \ding{174}. At that point, the ARUCO readings became
invalid. From that point onward, the robot relied only on the model
estimated so far. Without colliding with the workspace obstacles,
the robot reached $\protect\dq x_{d,1}$ with reasonable accuracy,
as shown in snapshot \ding{175}. During the entire motion, the estimated
model never violated a constraint. The real model seems to have slightly
penetrated a constraint (< 2 mm), but that amount is within ARUCO's
expected error margin. Moreover, the conservative nature of our VFI
specification prevented any physical collision.}}
}
\end{figure*}

The results of the experiment in terms of real task error $\realtaskerror$,
estimated task error $\estimatedtaskerror$, measurement error $\measurementerror$,
and minimum distance to obstacles are summarized in Fig.~\ref{fig:E2_results}.
The minimum distanceerror is the minimum distance obtained from all
36 collision pairs and a negative value indicates a penetration. Snapshot
\ding{172} stands for the beginning of the experiment, where the end-effector
was outside the box with a poorly estimated initial model. For 10
seconds, the parameters were updated using the ARUCO readings without
moving the robot. This caused a near convergence of the measurement
error and sharply improved the estimated model. The robot was then
moved while adapting the parameters until snapshot \ding{173}, where
the robot converged to the first setpoint $\dq x_{d,0}$. At that
point, all errors show convergent behavior and the setpoint was changed.
The robot used the high-quality model obtained so far to move to the
second setpoint, while continuously adapting the model until snapshot
\ding{174}. At that point, the ARUCO readings became invalid. From
that point onward, the robot relied only on the model estimated so
far. Without colliding with the workspace obstacles, the robot reached
$\dq x_{d,1}$ with reasonable accuracy, as shown in snapshot \ding{175}.
During the entire motion, the estimated model never violated a constraint.
This is specially important because only the violations of the \emph{estimated}
model are relevant to the feasibility of the controller, and by consequence,
to the closed-loop stability. The real model seems to have slightly
penetrated a constraint (< 2 mm), but that amount is within ARUCO's
expected error margin. Moreover, the conservative nature of our VFI
specification prevented any physical collision.

The results of the experiment in terms of computational time are summarized
in Table~\ref{tab:experiment_CA_computational_time}. The time spent
communicating with the robot, datalogging, etc. are abbreviated as
(\emph{Comm.}). In average, the time for computing $\myvec u_{q}$
using the task controller was about $0.1$~ms for a 6-DoF robot with
60 inequality constraints; the time for computing $\myvec u_{\hat{a}}$
using the adaptation controller was $7$~ms, considering 36 parameters
and 108 inequality constraints; and computations unrelated to the
controllers took $4.9$~ms (for instance, the one-way communication
with the robot takes about $2.0$~ms). All computations were done
within the sampling time of 20~ms, with great safety margin, in all
control modes.

These results indicate that the proposed strategy is an effective
way to mitigate errors in the nominal model when measurements are
available. Moreover, we have shown that when measurements are unavailable,
the robot can perform well with the model updated up to that point.
Our reproducible experimental setup can become a benchmark for future
works in the field of constrained adaptive kinematic control.

\textcolor{blue}{}
\begin{table}[tbh]
\textcolor{black}{\caption{\textcolor{black}{\label{tab:experiment_CA_computational_time}Computational
time for the experiment in Section~\ref{sec:E2_validation}.}}
}
\begin{centering}
\textcolor{blue}{}%
\begin{tabular}{|c|c|c|}
\hline 
 & Mean {[}ms{]} & Std {[}ms{]}\tabularnewline
\hline 
\hline 
Adaptive + Comm. ($t_{\mathrm{AC}}$) & 11.9 & 0.32\tabularnewline
Adaptive + Task + Comm. ($t_{\mathrm{ATC}}$) & 12.0 & 0.20\tabularnewline
Task + Comm. ($t_{\mathrm{TC}}$) & 5.0 & 0.24\tabularnewline
\hline 
Adaptive: $t_{\mathrm{ATC}}-t_{\mathrm{TC}}$ & 7 & -\tabularnewline
Task: $t_{\mathrm{ATC}}-t_{\mathrm{AC}}$ & 0.1 & -\tabularnewline
\hline 
\end{tabular}
\par\end{centering}
\textcolor{blue}{\medskip{}
}

The time spent communicating with the robot, datalogging, etc. are
abbreviated as (Comm.). In terms of average behavior, the task controller
took about $0.1$~ms for a 6-DoF robot with 60 inequality constraints;
the adaptation controller took $7$~ms, for 36 parameters and 108
inequality constraints; and computations unrelated to the controllers
took $4.9$~ms (for instance, the one-way communication with the
robot takes about $2.0$~ms). All computations were done within the
sampling time of 20~ms, with great safety margin, in all control
modes.
\end{table}

\textcolor{blue}{}

\section{Conclusions}

In this work, we proposed an adaptive constrained kinematic control
strategy for robots with arbitrary geometry. Our strategy is based
on solving two quadratic-programming problems to generate the instantaneous
control inputs. The first one, the task-space control law, reduces
the estimated task error and the second one, the adaptation law, reduces
the measurement error. We have shown that the closed-loop system under
those two laws is Lyapunov stable.

Experiments have shown that, even if starting from only a rough offline
calibration, our control strategy is effective when using a measurement
system in the sense that the error between the estimated end-effector
pose and the desired pose is always non-increasing, as our theoretical
analyses predict. Also, the end-effector measurement error and the
real task error always tend to improve, even when only partial measurements
are available, whereas all errors in the component related to the
measured variable also tend to improve. Errors in the components related
to the unmeasured parts of the task-space, however, may increase compared
to the baseline (i.e., when there is no task-space measurement and
the control is done using the nominal forward kinematics) even if
the overall errors in the task-space decrease. This is because even
though we ensure that the adaptation law does not change the task
vector in the direction of unmeasured components, the robot may execute
different trajectories when different components are measured. Therefore,
when compared to the baseline, it may be the case that the errors
along unmeasured components increase.

The experimental results also have shown that our control strategy
is capable of handling VFIs both in the task-space control law and
adaptation law, which enables the inclusion of nonlinear constraints
in the task-space variables as linear differential inequalities in
the control inputs. Therefore, we can ensure that geometrical constraints
in the workspace are respected while the robot parameters are adapted,
increasing the overall safety.

The combination of controller parameters, such as task-space error
gains and adaptation gains determine the ratio between task-space
convergence and adaptation. The VFI gains and the parameter $\alpha$
that is used to split the VFIs between the task-space and adaptive
control laws determine how fast the system is allowed to approach
constraints, such as obstacles and joint limits, while accounting
for the adaptation. Although they should be chosen in a way that prevents
the robot from moving too fast before the model is properly adapted,
the overall system behaves well for a wide range of values and combinations.
For instance, trivial choices such as $\alpha=0.5$, which means that
the VFI gains are the same for both task-space and adaptation control
laws, are suitable for real applications. Nonetheless, a more systematic
procedure for tuning the parameters will be investigated in the future.

Future works will also focus on adaptively changing the safety distances
of the VFIs according to the uncertainties associated with the measurement
error in the adaptation law to formally ensure that constraints are
not violated in the real task-space, without being overly conservative
(i.e, making safe distances much larger than necessary). Also, we
will extend our method to account for the second-order differential
kinematics and the full robot dynamics using the (constrained) Euler-Lagrange
equations.

\appendices{}

\section{Proof of Lemma~\ref{lem:distance}\label{sec:Additional-proofs}}

Since $\norm{\myvec t}=\norm{\myvec t_{n}}=R$ and $\norm{\estimated{\myvec t}}\geq R$,
then $\norm{\estimated{\myvec t}}=\beta R$ and $\norm{\myvec t_{\lambda}}=\alpha R$
with $\beta\geq\alpha\geq1$. Hence, $\myvec t_{n}=\estimated{\myvec t}/\beta$
and $\myvec t_{\lambda}=\estimated{\myvec t}\left(\beta-\beta\lambda+\lambda\right)/\beta$.
Thus, $\norm{\myvec t_{\lambda}}=\left(\beta-\beta\lambda+\lambda\right)R$,
which implies $\alpha=\beta-\beta\lambda+\lambda$ and $\myvec t_{\lambda}=\estimated{\myvec t}\left(\alpha/\beta\right)$.

Let $D_{\estimated tt}\triangleq\norm{\estimated{\myvec t}-\myvec t}^{2}$
and $D_{\lambda t}=\norm{\myvec t_{\lambda}-\myvec t}^{2}$. Since
$D_{\estimated tt}\geq D_{\lambda t}$ implies $\norm{\estimated{\myvec t}-\myvec t}\geq\norm{\myvec t_{\lambda}-\myvec t}$,
it suffices to show that $D_{\estimated tt}\geq D_{\lambda t}$. Let
us assume, for the sake of contradiction, that $D_{\estimated tt}<D_{\lambda t}$.
Since $D_{\estimated tt}=\norm{\estimated{\myvec t}}^{2}-2\myvec t^{T}\estimated{\myvec t}+\norm{\myvec t}^{2}$
and $D_{\lambda t}=\norm{\myvec t_{\lambda}}^{2}-2\myvec t^{T}\myvec t_{\lambda}+\norm{\myvec t}^{2}$,
then $(\beta R)^{2}-2\myvec t^{T}\estimated{\myvec t}+R^{2}<(\alpha R)^{2}-2\myvec t^{T}\estimated{\myvec t}\left(\alpha/\beta\right)+R^{2}$,
which implies $(\beta^{2}-\alpha^{2})R^{2}<2\myvec t^{T}\estimated{\myvec t}\left[(\beta-\alpha)/\beta\right]$
and thus $(\beta+\alpha)R^{2}<2\myvec t^{T}\estimated{\myvec t}/\beta$.
Because $\myvec t^{T}\estimated{\myvec t}=\norm{\myvec t}\norm{\estimated{\myvec t}}\cos\phi_{t\estimated t}$,
we obtain $\beta+\alpha<2\cos\phi_{t\estimated t}$. Since $(2\cos\phi_{t\estimated t})\in[-2,2]$,
we conclude that $\beta+\alpha<2$, which is a contradiction because
$\beta+\alpha\geq2$. Therefore, it is not true that $D_{\estimated tt}<D_{\lambda t}$.
Hence, $D_{\estimated tt}\geq D_{\lambda t}$, which concludes the
proof.

\section{Parameterization of rigid motions, rotations, and translations\label{sec:Implementation}}

The use cases explored in this work have been defined in general terms
in Section~\ref{sec:Use-cases}. Nonetheless, their implementation
depends on the choice of parameterization. We use dual quaternion
algebra to represent rigid motion transformations, in which translations
and rotations are particular cases. Since our general formulation
is not dependent on any particular parameterization, here we do not
present any comparison of dual quaternion algebra with other representations.
Instead, the goal of this appendix is to ensure completeness of presentation
and reproducibility in case readers want to replicate our results,
which would require using the same representation that we have used.
Interested readers can find a gentle introduction to dual quaternion
algebra in \cite{adorno2017robot}. A discussion about its advantages
in the context of constrained control can be found in \cite{Marinho2019}.
The computational library that we use to manipulate elements of dual
quaternion algebra is described in \cite{adorno2020dqrobotics}.

Let the quaternion set be
\begin{align*}
\mathbb{H} & \triangleq\left\{ h_{1}+\imi h_{2}+\imj h_{3}+\imk h_{4}\,:\,h_{1},h_{2},h_{3},h_{4}\in\mathbb{R}\right\} 
\end{align*}
with $\hat{\imath}^{2}=\hat{\jmath}^{2}=\hat{k}^{2}=\hat{\imath}\hat{\jmath}\hat{k}=-1$
and the dual quaternion set be
\[
\mathcal{H}\triangleq\left\{ \quat h+\dual\quat h':\quat h,\quat h'\in\mathbb{H},\,\dual^{2}=0,\,\dual\neq0\right\} .
\]

\subsection{Positions and translations\label{subsec:Positions-and-translations}}

The set of pure quaternions is given by $\mathbb{H}_{p}\triangleq\left\{ \quat h\in\mathbb{H}\,:\,\real{\quat h}=0\right\} $,
where $\real{h_{1}+\imi h_{2}+\imj h_{3}+\imk h_{4}}=h_{1}$, and
is isomorphic to $\mathbb{R}^{3}$ under the addition operation.

\subsection{Orientations and rotations\label{subsec:Orientations-and-rotations}}

The unit quaternion set, $\mathbb{S}^{3}\triangleq\left\{ \quat h\in\mathbb{H}\,:\,\norm{\quat h}=1\right\} $,
contains elements that represent orientations and rotations in the
tridimensional space. When the set $\mathbb{S}^{3}$ is equipped with
the multiplication operation, we obtain the group $\spin$ of rotations,
which double covers $\mathrm{SO}(3)$.

\subsection{Poses and rigid motions\label{subsec:Poses-and-rigid}}

Elements of the unit dual quaternion set, $\dq{\mathcal{S}}\triangleq\left\{ \quat r+\left(1/2\right)\dual\quat t\quat r:\quat r\in\mathbb{S}^{3},\quat t\in\mathbb{H}_{p}\right\} \subset\mathcal{H}$,
represent poses and rigid motions in the tridimensional space. Analogously
to $\spin$, when $\dq{\mathcal{S}}$ is equipped with the multiplication
operation, we obtain the group $\spinr$, which double covers $\mathrm{SE}(3)$.

\subsection{Error definition\label{subsec:Error-definition}}

\begin{table}[tbh]
\caption{\label{tab:complete_and_partial_values}Complete and partial task-space
values.}

\noindent\resizebox{1.0\columnwidth}{!}{%
\begin{centering}
\begin{tabular}{cc}
\hline 
 & Estimated\tabularnewline
\hline 
\hline 
$\dq y,\quat y_{r},\quat y_{t},y_{d}$ & Measured pose, rotation, translation, and distance, respectively.\tabularnewline
$\dq x_{d},\quat r_{d},\quat t_{d},d_{d}$ & Desired pose, rotation, translation, and distance, respectively.\tabularnewline
$\estimated{\dq x},\estimated{\quat r},\estimated{\quat t},\estimated d$ & Estimated pose, rotation, translation, and distance, respectively.\tabularnewline
$\dq x,\quat r,\quat t,d$ & Real pose, rotation, translation, and distance, respectively.\tabularnewline
\hline 
\end{tabular}
\par\end{centering}
}
\end{table}

\begin{table}[tbh]
\caption{\label{tab:complete_and_partial_errors}Complete and partial error
definitions.}

\noindent\resizebox{1.0\columnwidth}{!}{%
\begin{centering}
\begin{tabular}{cccc}
\hline 
 & Estimated & Real & Measurement\tabularnewline
\hline 
\hline 
Distance & $\breve{d}=\estimated d-d_{d}$ & $\tilde{d}=d-d_{d}$ & $\tilde{y}_{d}=\estimated d-y_{d}$\tabularnewline
Translation & $\breve{\quat t}=\estimated{\quat t}-\quat t_{d}$ & $\tilde{\quat t}=\quat t-\quat t_{d}$ & $\tilde{\quat y}_{t}=\estimated{\quat t}-\quat y_{t}$\tabularnewline
Rotation & $\breve{\quat r}=\mathcal{E}\left(\estimated{\quat r},\quat r_{d}\right)$ & $\tilde{\quat r}=\mathcal{E}\left(\quat r,\quat r_{d}\right)$ & $\tilde{\quat y}_{r}=\mathcal{E}\left(\estimated{\quat r},\quat y_{r}\right)$\tabularnewline
Pose & $\breve{\dq x}\triangleq\mathcal{E}\left(\estimated{\dq x},\dq x_{d}\right)$ & $\tilde{\dq x}\triangleq\mathcal{E}\left(\dq x,\dq x_{d}\right)$ & $\tilde{\dq y}\triangleq\mathcal{E}\left(\estimated{\dq x},\dq y\right)$\tabularnewline
\hline 
\end{tabular}
\par\end{centering}
}
\end{table}

Consider a desired end-effector pose $\dq{\mathcal{S}}\ni\dq x_{d}=\quat r_{d}+\dual\frac{1}{2}\quat t_{d}\quat r_{d}$,
in which $\quat r_{d}\in\mathbb{S}^{3}$ is the desired end-effector
orientation and $\quat t_{d}\in\mathbb{H}_{p}$ is the desired end-effector
position. Analogously, let $\estimated{\dq x}\in\dq{\mathcal{S}}$
be the estimated end-effector pose, $\dq x\in\dq{\mathcal{S}}$ be
the real end-effector pose, and $\dq y\in\dq{\mathcal{S}}$ be the
measured end-effector pose. These elements are summarized in Table~\ref{tab:complete_and_partial_values}.

Using the aforementioned elements, the error definitions are summarized
in Table~\ref{tab:complete_and_partial_errors} and further explained
as follows. The estimated distance error and translation error, similarly
to their counterparts parameterized with elements of the Euclidean
group $\left(\mathbb{R}^{3},+\right)$, are defined as $\breve{d}=\estimated d-d_{d}=\norm{\estimated{\quat t}}_{2}-\norm{\quat t_{d}}_{2}$
and $\breve{\quat t}=\estimated{\quat t}-\quat t_{d}$, respectively.
On the other hand, orientation and pose errors that respect the topology
of the underlying space of orientations and rigid motions are defined
by multiplications in both $\spin$ and $\spinr$, respectively \cite{Pham2017}.
For instance, the estimated pose error is defined as $\breve{\dq x}\triangleq\mathcal{E}\left(\estimated{\dq x},\dq x_{d}\right)$
such that
\begin{equation}
\mathcal{E}\left(\estimated{\dq x},\dq x_{d}\right)=\begin{cases}
\estimated{\dq x}^{*}\dq x_{d}-1 & \text{if }\norm{\estimated{\dq x}^{*}\dq x_{d}-1}_{2}<\norm{\estimated{\dq x}^{*}\dq x_{d}+1}_{2}\\
\estimated{\dq x}^{*}\dq x_{d}+1 & \text{otherwise},
\end{cases}\label{eq:invariant error-1}
\end{equation}
and the estimated rotation error as $\breve{\quat r}=\mathcal{E}\left(\estimated{\quat r},\quat r_{d}\right)$
\cite{Marinho2019a}, where $\estimated{\dq x}^{*}$ and $\estimated{\quat r}^{*}$
are the dual quaternion conjugate of $\estimated{\dq x}$ and quaternion
conjugate of $\estimated{\quat r}$, respectively. The reason for
using such pose and rotation errors is to prevent the problem of unwinding
\cite{Kussaba2017}, because $\breve{\dq x}$ and $-\breve{\dq x}$
represent the same pose, and similarly $\breve{\quat r}$ and $-\breve{\quat r}$
represent the same orientation.

\subsection{Mapping errors to vectors}

\noindent Using the bijective operators $\vector_{3}:\mathbb{H}_{p}\to\mathbb{R}^{3}$,
$\vector_{4}:\mathbb{H}\to\mathbb{R}^{4}$, and $\vector_{8}:\dq{\mathcal{S}}\to\mathbb{R}^{8}$,
which take the coefficients of (dual) quaternions and stack them into
a vector, we map those errors to vectors to comply with the formulation
in Sections~\ref{sec:Problem-Definition}–\ref{sec:Use-cases}.

\subsection{Error norms}

\noindent Rewriting the estimated pose error as $\breve{\dq x}=\mathcal{E}\left(\estimated{\dq x},\dq x_{d}\right)=\breve{\quat x}_{P}+\dual\breve{\quat x}_{D}$,
the estimated pose error norm used in Sections~\ref{sec:E1_validation}–\ref{sec:E2_validation}
is defined as $\norm{\breve{\dq x}}_{2}\triangleq\norm{\breve{\quat x}_{P}}_{2}+\norm{\breve{\quat x}_{D}}_{2}$,
whereas the orientation, position, and distance norms are defined
as $\norm{\breve{\quat r}}_{2}$, $\norm{\breve{\quat t}}_{2}$, and
$\norm{\breve{d}}_{2}=\left|\breve{d}\right|$, respectively.\footnote{The quaternion norm is equivalent to the Euclidean norm, but the dual
quaternion norm is not. Therefore, if $\quat h\in\mathbb{H}$, then
$\norm{\quat h}=\norm{\quat h}_{2}$. In contrast, given $\dq h\in\mathcal{H}$,
except for particular cases, usually $\norm{\dq h}\neq\norm{\dq h}_{2}$.}

\section{Analysis of closed-loop stability for multiplicative errors\label{sec:Appendix-Stability}}

Closed-loop stability is also guaranteed when using multiplicative
errors such as \eqref{eq:invariant error-1}. This can be shown by
following the procedure described in Section~\ref{subsec:Lyapunov-stability}.

Indeed, let the task-space be defined as the set of all end-effector's
poses, parameterized using $\dq{\mathcal{S}}$. Consider the Lyapunov
function $V(\estimatedtaskerror)=\frac{1}{2}\estimatedtaskerror^{T}\estimatedtaskerror$,
where
\begin{align}
\estimatedtaskerror & =\vector_{8}\left(\mathcal{E}\left(\estimated{\dq x},\dq x_{d}\right)\right).\label{eq:invariant-estimated-task-error}
\end{align}
The time derivative of $\estimatedtaskerror$ is given by $\dot{\estimatedtaskerror}=\vector_{8}\left(\dot{\estimated{\dq x}}^{*}\dq x_{d}\right)$
because $\dot{\dq x}_{d}=0$ for all $t$. Using the Hamilton operator
$\hami -_{8}:\mathcal{H}\to\mathbb{R}^{8\times8}$, such that $\vector_{8}\left(\dq a\dq b\right)=\hami -_{8}\left(\dq b\right)\vector_{8}\dq a$,
with $\dq a,\dq b\in\mathcal{H}$, we obtain $\dot{\estimatedtaskerror}=\hami -_{8}\left(\dq x_{d}\right)\mymatrix C_{8}\vector_{8}\dot{\estimated{\dq x}}$,
where $\mymatrix C_{8}\in\mathbb{R}^{8\times8}$ is the matrix that
satisfies $\vector_{8}\dq a^{*}=\mymatrix C_{8}\vector_{8}\dq a$
for all $\dq a\in\mathcal{H}$ \cite{adorno2017robot}. Defining $\dot{\hat{\myvec x}}\triangleq\vector_{8}\dot{\estimated{\dq x}}$,
we obtain $\dot{\breve{\myvec x}}=\hami -_{8}\left(\dq x_{d}\right)\mymatrix C_{8}\mymatrix J_{\hat{x}}\dot{\myvec v}$,
where $\mymatrix J_{\hat{x}}$ and $\dot{\myvec v}$ are defined as
in \eqref{eq:equivalent task-space control law}. Let $\text{\ensuremath{\mymatrix G}}_{\hat{x}}\triangleq\hami -_{8}\left(\dq x_{d}\right)\mymatrix C_{8}\mymatrix J_{\hat{x}}$,
then $\dot{\breve{\myvec x}}=\mymatrix G_{\hat{x}}\dot{\myvec v}$
and
\begin{align}
\dot{V}(\estimatedtaskerror) & =\estimatedtaskerror^{T}\mymatrix G_{\hat{x}}\dot{\myvec v}=\estimatedtaskerror^{T}\mymatrix G_{\hat{x}}\myvec u_{v,q}+\estimatedtaskerror^{T}\mymatrix G_{\hat{x}}\myvec u_{v,\estimated a}.\label{eq:derivative-lyapunov-function-invariant}
\end{align}
We partition $\text{\ensuremath{\mymatrix G}}_{\hat{x}}$ such that
$\mymatrix G_{\hat{x}}=\begin{bmatrix}\mymatrix G_{\hat{x},q} & \mymatrix G_{\hat{x},\hat{a}}\end{bmatrix}$
and replace $\mymatrix J_{\hat{x},q}$ with $\mymatrix G_{\hat{x},q}$
in \eqref{eq:problem_quadratic_constrained}, where $\estimatedtaskerror$
is defined in \eqref{eq:invariant-estimated-task-error}.

Now, consider two cases for the measurement space, where the end-effector
pose space and the end-effector orientation space are parameterized
as $\dq{\mathcal{S}}$ and $\mathbb{S}^{3}$, respectively. We define
$\tilde{\myvec y}=\vector_{8}\left(\mathcal{E}\left(\estimated{\dq x},\dq x\right)\right)$
for the pose or $\tilde{\myvec y}=\vector_{4}\left(\mathcal{E}\left(\estimated{\quat r},\quat r\right)\right)$
for the orientation.

Using any of those definitions, we replace $\mymatrix J_{\estimated y,\estimated a}$
with $\mymatrix G_{\estimated y,\estimated a}$ and $\mymatrix J_{\estimated x,\estimated a}$
with $\mymatrix G_{\estimated x,\estimated a}$ in \eqref{eq:adaptive-constrained-law},
where $\mymatrix G_{\hat{y}}=\begin{bmatrix}\mymatrix G_{\hat{y},q} & \mymatrix G_{\hat{y},\hat{a}}\end{bmatrix}=\hami -_{8}\left(\dq x\right)\mymatrix C_{8}\mymatrix J_{\hat{x}}$
for the pose or $\mymatrix G_{\hat{y}}=\hami -_{4}\left(\quat r\right)\mymatrix C_{4}\mymatrix J_{\estimated{\quat r}}$
for the orientation, where $\hami -_{4}:\mathbb{H}\to\mathbb{R}^{4\times4}$
satisfies $\vector_{4}(\quat a\quat b)=\hami -_{4}(\quat b)\vector_{4}\quat a$,
with $\quat a,\quat b\in\mathbb{H}$, and $\mymatrix C_{4}$ satisfy
$\vector\quat r^{*}=\mymatrix C_{4}\vector_{4}\quat r$.

Therefore, by replacing $\mymatrix J$ with $\mymatrix G$ (using
the appropriate subscripts), the analysis is essentially the same,
and thus we conclude that multiplicative errors such as \eqref{eq:invariant-estimated-task-error}
do not affect closed-loop stability.

\bibliographystyle{IEEEtran}
\bibliography{bib/ral}

\begin{IEEEbiography}[{\includegraphics[width=1in,height=1.25in]{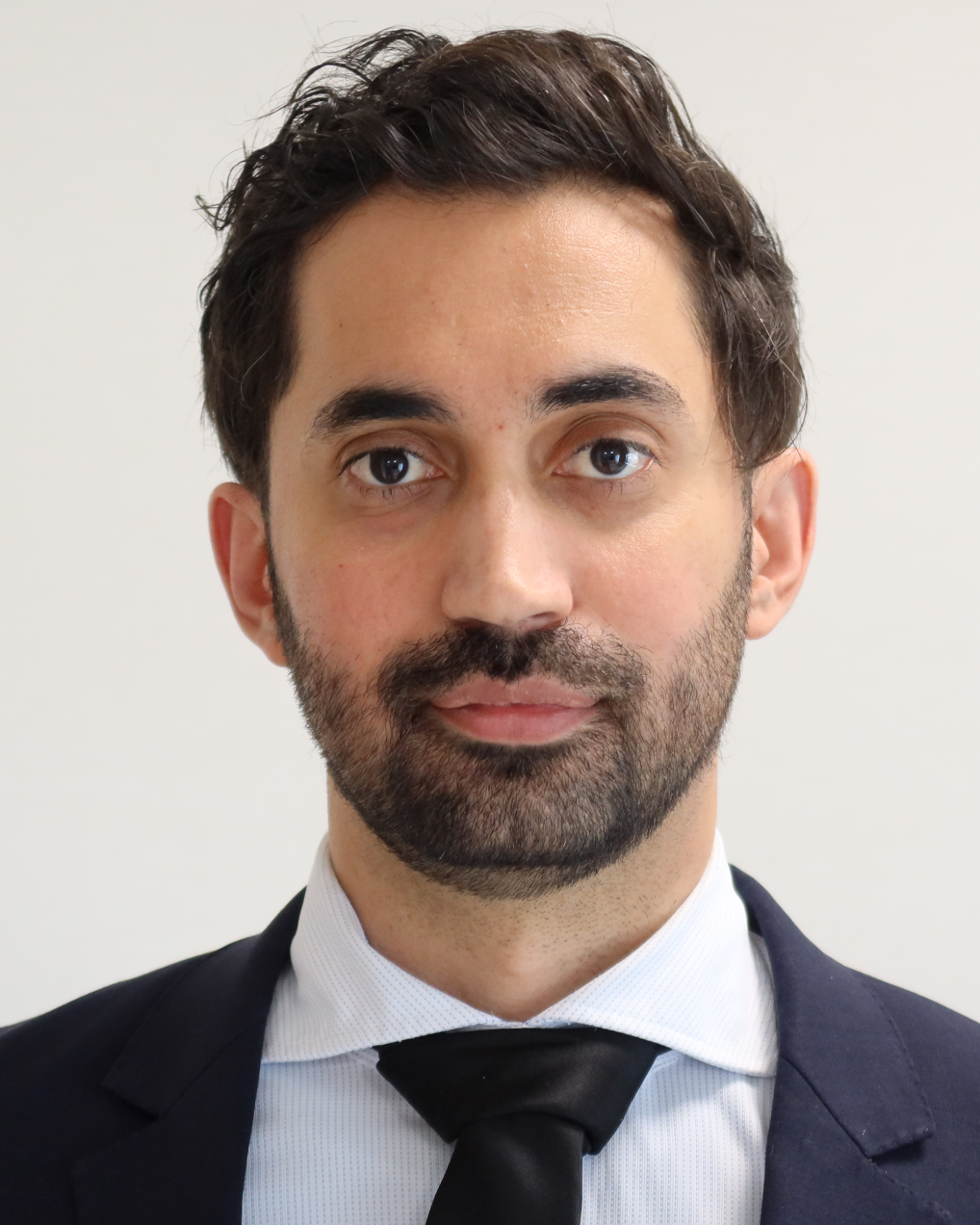}}]{Murilo Marques Marinho}
 (GS’13–M'18) received the bachelor's degree in mechatronics engineering
and the master's degree in electronic systems and automation engineering
from the University of Brasilia, Brasilia, Brazil, in 2012 and 2014,
respectively. He received the Ph.D. degree in mechanical engineering
from the University of Tokyo, Tokyo, Japan, in 2018. In 2018, he was
a Visiting Researcher with the Johns Hopkins University. He is an
Assistant Professor with the University of Tokyo from 2019. His research
interests include robotics applied to medicine, robot control theory,
and image processing.
\end{IEEEbiography}

\begin{IEEEbiography}[{\includegraphics[clip,width=1in,height=1.25in]{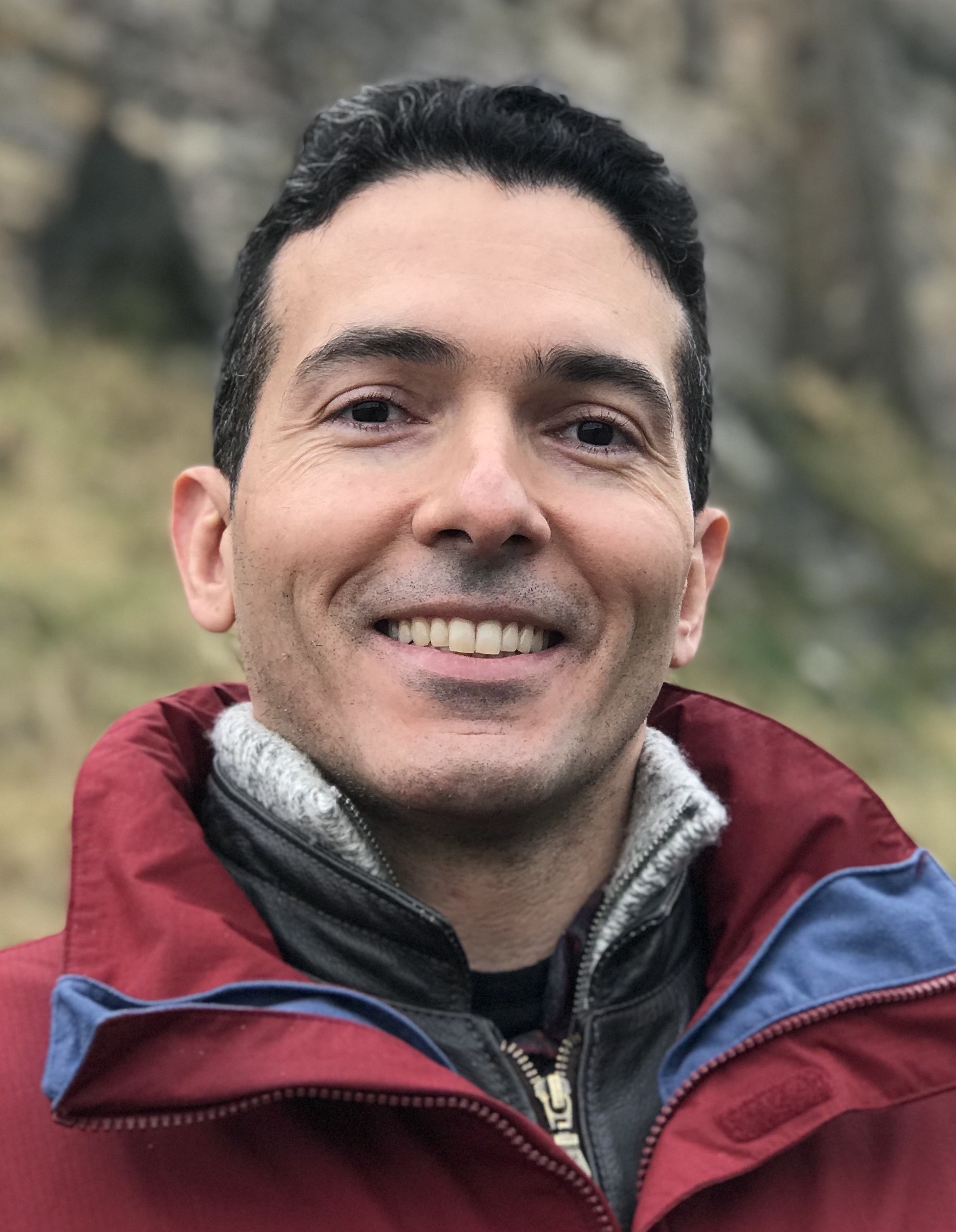}}]{Bruno Vilhena Adorno}
 (GS’09–M’12–SM’17) received the BSc degree in mechatronics engineering
and the MSc degree in electrical engineering from the University of
Brasilia, Brasilia, Brazil, in 2005 and 2008, respectively, and the
Ph.D. degree in automatic and microelectronic systems from the University
of Montpellier, Montpellier, France, in 2011. He is currently a Senior
Lecturer in Robotics with the Department of Electrical and Electronic
Engineering at The University of Manchester (UoM), Manchester, United
Kingdom. Before joining the UoM, he was an Associate Professor with
the Department of Electrical Engineering at the Federal University
of Minas Gerais, Belo Horizonte, Brazil. His current research interests
include both practical and theoretical aspects related to robot kinematics,
dynamics, and control with applications to mobile manipulators, humanoids,
and cooperative manipulation systems.
\end{IEEEbiography}

\end{document}